\pdfoutput=1

\documentclass[11pt]{article}

\usepackage[preprint]{acl}

\usepackage{times}
\usepackage{latexsym}

\usepackage[skip=4pt]{caption}
\usepackage[T1]{fontenc}

\usepackage[utf8]{inputenc}
\usepackage{multirow}
\usepackage{microtype}
\usepackage{makecell}
\usepackage{inconsolata}

\usepackage{amsmath}
\usepackage{graphicx}
\usepackage{booktabs}
\usepackage{makecell}
\usepackage{siunitx}
\usepackage[table]{xcolor}
\definecolor{rowgray}{gray}{0.95}
\usepackage{caption}
\usepackage{tcolorbox}
\usepackage{algorithm}
\usepackage[noend]{algpseudocode}

\title{Rethinking Schema Linking: A Context-Aware Bidirectional Retrieval Approach for Text-to-SQL}

\author{
Md Mahadi Hasan Nahid$^{\clubsuit}$ \thanks{This work was conducted during an internship at \emph{Huawei Technologies Canada Co., Ltd., Burnaby, Canada}.} \enspace Davood Rafiei$^{\clubsuit}$ \enspace Weiwei Zhang$^{\diamondsuit}$ \enspace Yong Zhang$^{\diamondsuit}$\\
$^{\clubsuit}$University of Alberta, Canada\\
$^{\diamondsuit}$Huawei Technologies Canada Co., Ltd., Canada\\
$^{\clubsuit}$\texttt{\{mnahid, drafiei\}@ualberta.ca} \enspace $^{\diamondsuit}$\texttt{\{weiwei.zhang2, yong.zhang3\}@huawei.com} \\
}

\begin{document}

\maketitle
\begin{abstract}
Schema linking---the process of aligning natural language questions with database schema elements---is a critical yet underexplored component of Text-to-SQL systems. While recent methods have focused primarily on improving SQL generation, they often neglect the retrieval of relevant schema elements, which can lead to hallucinations and execution failures. In this work, we propose a context-aware bidirectional schema retrieval framework that treats schema linking as a standalone problem. Our approach combines two complementary strategies: table-first retrieval followed by column selection, and column-first retrieval followed by table selection. It is further augmented with techniques such as question decomposition, keyword extraction, and keyphrase extraction. Through comprehensive evaluations on challenging benchmarks such as BIRD and Spider, we demonstrate that our method significantly improves schema recall while reducing false positives. Moreover, SQL generation using our retrieved schema consistently outperforms full-schema baselines and closely approaches oracle performance, all without requiring query refinement. Notably, our method narrows the performance gap between full and perfect schema settings by 50\%. Our findings highlight schema linking as a powerful lever for enhancing Text-to-SQL accuracy and efficiency.

\end{abstract}

\section{Introduction}

Text-to-SQL (NL2SQL) systems translate natural language questions into executable SQL queries, enabling non-expert users to interact with relational databases. Despite progress driven by large language models (LLMs), a critical bottleneck remains: accurately identifying which parts of the database schema are relevant to a given question. This step, known as \textit{schema linking}, is foundational to the NL2SQL pipeline but is often treated as a secondary concern. 

\begin{figure*}[ht]
    \centering
    \resizebox{0.90\textwidth}{!}{
    \includegraphics{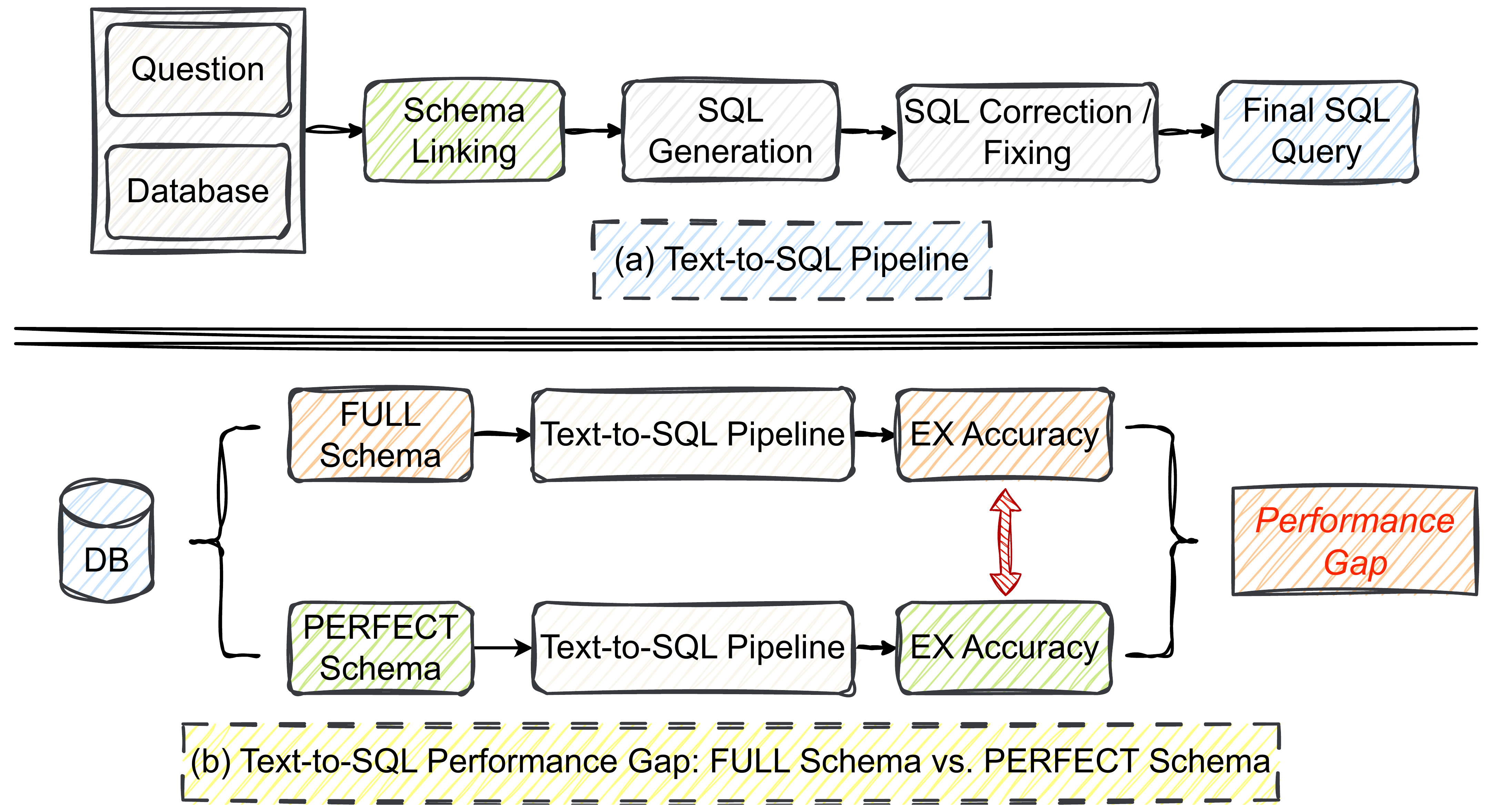}
    }
    \caption{Overview of the Text-to-SQL pipeline highlighting the performance gap between using a perfect schema and the gold schema. The figure demonstrates how schema linking plays a crucial role in narrowing this performance gap by retrieving a schema close to the perfect schema, which significantly improves the SQL query generation process.}
    \label{fig:performance-gap}
\end{figure*}

Using the full schema during SQL generation introduces irrelevant context, increases token overhead, and often leads to hallucinations---factors that collectively reduce execution accuracy and establish a practical lower bound on performance (see Figure~\ref{fig:performance-gap}) \cite{liu-etal-2024-lost, laban-etal-2024-summary}.
In contrast, using a \textit{perfect} schema---one that includes only the necessary tables and columns---yields significantly higher accuracy and serves as an upper bound.
While some existing approaches attempt to close this gap through query refinement or post-generation filtering, few tackle schema linking directly or strive for high-recall retrieval that is both accurate and efficient~\cite{pourreza2024din, talaei2024chesscontextualharnessingefficient, lee-etal-2025-mcs}.
For example, CHESS \cite{talaei2024chesscontextualharnessingefficient} achieves strong results but depends on multiple LLM calls for schema retrieval, making it computationally expensive \cite{chung2025longcontextneedleveraging}.
The question studied in this paper is if the perfect schema can be closely approximated by efficiently retrieving a relevant subset of the schema that 
achieves high recall and low false positive rate—without relying on downstream SQL correction or refinement.

To this end, we treat schema linking as an independent and central problem in the NL2SQL pipeline. Our approach introduces a bidirectional schema retrieval framework that integrates two complementary strategies: table-first retrieval followed by column selection, and column-first retrieval followed by table selection. To further enhance precision and recall, we incorporate augmentation techniques such as question decomposition, keyword extraction, and value identification---tools previously shown to improve SQL generation \cite{pourreza2024din, talaei2024chesscontextualharnessingefficient, lee-etal-2025-mcs}. We hypothesize that these same techniques can also guide LLMs toward retrieving more accurate and compact schema representations.

\vspace{-0.7em}

Our extensive evaluations on state-of-the-art Text-to-SQL benchmarks demonstrate that our method consistently improves schema recall while minimizing irrelevant context. Moreover, SQL generation using our retrieved schema outperforms full-schema baselines and approaches oracle performance without requiring post-hoc query refinement.


Our contributions can be summarized as follows:
\textbf{(1)} We propose a novel bidirectional schema retrieval framework that integrates table-first and column-first strategies to improve schema linking recall while reducing false positives.
\textbf{(2)} We enhance schema linking using augmentation techniques such as question decomposition, keyword extraction, and condition parsing to improve contextual grounding and retrieval precision.
\textbf{(3)} We demonstrate that treating schema linking as a standalone task leads to significant gains in SQL generation accuracy and efficiency, even in zero-shot settings without downstream query refinement.
\textbf{(4)} We conduct extensive experiments on challenging datasets such as BIRD, showing that our retrieved schema outperforms full-schema baselines and closely approximates oracle-level performance with reduced computational overhead.

\begin{figure*}[ht]
    \centering
    \resizebox{1.0\textwidth}{!}{
    \includegraphics{figures/bi-sl-v2-sketch-3.pdf}
    }
    \caption{Illustration of our proposed Bidirectional Schema Linking Approach. The diagram provides a detailed overview of how our method effectively refines schema linking by leveraging both table-first and column-first strategies, resulting in more accurate schema retrieval and enhanced performance in Text-to-SQL tasks.}
    \label{fig:method-sl-bi}
\end{figure*}

\section{Related Work}

\paragraph{Text-to-SQL Pipelines.}
Earlier Text-to-SQL models adopted modular pipelines, separating schema linking, parsing, and SQL generation. For example, IRNet used intermediate representations (SemQL) to bridge NL queries and SQL forms \cite{liu2024surveynl2sqllargelanguage}. Lei et al. \cite{lei-etal-2020-examining} highlighted the crucial role of schema linking by providing a labeled corpus and showing that performance heavily depends on accurate linking.

Several works enhanced traditional pipelines through query decomposition and reasoning~\cite{dong2023c3zeroshottexttosqlchatgpt, pourreza2024din, talaei2024chesscontextualharnessingefficient}. DIN-SQL \cite{pourreza2024din} introduced self-correction and step-by-step reasoning with decomposed in-context learning. CHASE-SQL \cite{pourreza2024chasesqlmultipathreasoningpreference} further extended this idea by combining multi-path reasoning and candidate reranking. These works laid the groundwork for understanding the interplay between schema linking and accurate SQL generation.

\paragraph{Schema Linking in Text-to-SQL.}

Early models such as RAT-SQL incorporated schema linking into the encoder using relation-aware self-attention to model interactions between questions and schema components~\cite{wang-etal-2020-rat}. SLSQL~\cite{lei-etal-2020-examining} demonstrated that schema linking alone, when done effectively using contextualized encoders, can yield strong results without complex decoding pipelines. However, traditional methods often relied on exact string matching, which is brittle against paraphrasing and linguistic variation. To address this, semantic-enhanced approaches such as ISESL-SQL~\cite{huang2022isesl} construct schema-linking graphs capturing deeper relationships between question tokens and schema elements.

The emergence of large language models (LLMs) has reignited debate on the necessity of explicit schema linking. Maamari et al.~\cite{maamari2024the} argue that well-reasoned LLMs may implicitly perform schema linking, especially when the full schema fits within the model’s context window. While promising, this often leads to increased computational costs and a greater risk of including irrelevant schema elements, which in turn degrade query generation accuracy~\cite{li2024can, maamari2024the, cao2024rsl}.  

To mitigate these issues, several recent works propose schema pruning techniques. SQL-to-Schema~\cite{10.1007/978-3-031-68309-1_11} and RSL-SQL~\cite{cao2024rsl} generates SQL over the full schema, then backtraces required tables and columns to construct a reduced schema context. Similarly, E-SQL~\cite{caferoğlu2025esqldirectschemalinking} enhances the original question with schema elements, helping align user intent with the database structure. \citet{kothyari-etal-2023-crush4sql} introduces a retrieval-based method using hallucinated schema elements and candidate reranking, tightly coupled with SQL generation. The recent DTS-SQL \cite{pourreza-rafiei-2024-dts} framework shows that separating schema linking and SQL generation through a two-stage fine-tuning process enhances the performance of smaller models, underscoring the critical role of schema linking. \citet{gorti-etal-2025-msc} propose MSc-SQL, which critiques multiple sampled SQL outputs using execution feedback and metadata to select the best candidate, enabling small open-source models to rival larger closed ones.


\vspace{-0.5em}
\paragraph{Augmentation and Robust Linking.}

Other strategies focus on context augmentation. RSL-SQL~\cite{cao2024rsl} improves linking robustness by combining bidirectional attention and partial query execution, enabling the model to validate schema selections. Solid-SQL~\cite{liu-etal-2025-solid} employs structured similarity-based in-context retrieval to provide stronger schema grounding during inference. TA-SQL~\cite{qu-etal-2024-generation} leverages Task Alignment to reduce hallucinations in schema selection and text-to-SQL generation, encouraging LLMs to draw on similar tasks and thereby improving robustness and reliability.

Despite these advancements, many of these methods either tightly couple schema linking with SQL generation or require extensive LLM calls, as in CHESS~\cite{talaei2024chesscontextualharnessingefficient}, which reduces interpretability and scalability \cite{chung2025longcontextneedleveraging, dong2023c3zeroshottexttosqlchatgpt}.



\section{Methodology}

Our approach consists of two primary stages: Question Augmentation and Bidirectional Schema Linking. 
A visual representation of the overview of our proposed schema linking approach is provided in \textbf{Figure \ref{fig:method-sl-bi}}. 

\subsection{Question Augmentation}
\label{sec:QuestioAugmentation}
The initial stage involves enriching the natural language query through a process we refer to as \emph{Question Augmentation}. In this stage, the original query and hints are provided to an LLM under a set of carefully designed instructions. The model is tasked with extracting keys elements that are essential for accurate SQL generation. 
These elements include \emph{keywords and key phrases} that help identify relevant tables and columns, as well as filtering cues that inform the construction of WHERE clauses. The model also performs \emph{query decomposition}, breaking down complex questions into simpler subcomponents to reduce ambiguity and improve alignment with the database schema.


This augmented information encapsulates the key components required for effective schema linking, and appending it to the original query allows the LLM to more accurately contextualize and interpret the underlying database schema.

\subsection{Bidirectional Schema Linking}

While schema linking is often treated as a one-shot filtering step, we introduce a bidirectional strategy that decomposes the retrieval process into two complementary perspectives: \textit{table-first} and \textit{column-first} schema linking. Natural language questions convey cues that sometimes align more clearly with tables (representing entities) and at other times with columns (representing attributes). This dual-path design enables the model to identify schema elements from both high-level (table) and fine-grained (column) perspectives. Importantly, both procedures are guided by the same enriched query information generated during the Question Augmentation phase. This ensures consistency between schema retrieval and subsequent SQL generation, as the augmented elements keywords, conditions, and subquestions can be reused throughout the pipeline. Each pathway follows a two-stage refinement process using LLM calls, which reduces reliance on over-specified or hallucinated context. Below, we describe both directions in detail.

\subsubsection{Table-First Retrieval}

This approach begins by identifying relevant tables and subsequently narrowing down to specific columns:

\paragraph{(1) Table Filtering.}
The LLM receives the full schema, distinct sample column values, original query, and its augmented version. With carefully crafted instructions, it selects the subset of tables that are semantically aligned with the question's intent and constraints. This early pruning helps remove irrelevant relations, which can otherwise confuse SQL generation models.
\paragraph{(2) Column Selection.}
Using the filtered schema from the previous step, the LLM identifies relevant columns from the retained tables. This refinement phase ensures that only those attributes necessary for SELECT, WHERE, and GROUP BY clauses are retained. It reduces schema noise while preserving recall.

\subsubsection{Column-First Retrieval}

The reverse pathway begins at the attribute level and uses that context to derive table-level grounding:

\paragraph{(1) Column Filtering.}
The LLM is prompted to isolate columns from the schema that directly align with the key phrases and decomposed subquestions from the augmentation phase. This step is particularly effective when user queries are attribute-centric (e.g., asking for "average enrollment" or "admission rate").
\paragraph{(2) Table Selection.}
Given the set of selected columns, the LLM determines the tables that contain these columns. This ensures the contextual linkage between columns and their parent tables is respected, even in normalized or multi-table databases.

\subsection{Schema Merging}

The final stage of our framework combines the outputs from both the table-first and column-first schema linking pathways. Each direction captures complementary aspects of schema relevance—one from a structural perspective, the other from a fine-grained attribute perspective. To synthesize these strengths, we apply a schema merging operation, which performs a union over the tables selected in both directions and aggregates the corresponding relevant columns.

Rather than choosing a single path or arbitrarily prioritizing one over the other, this merging strategy is designed to maximize recall without compromising precision. Specifically:

\begin{itemize} 
    \item \textbf{Table Union:} We merge the set of relevant tables identified by both pathways. If a table appears in either direction, it is included in the final schema, reflecting its potential utility in the SQL query. 
    
    \item \textbf{Column Aggregation:} For each selected table, we unify the columns retrieved by both the table-first and column-first directions. This ensures no essential column is excluded while minimizing inclusion of irrelevant fields. 
\end{itemize}

This hybrid merging yields a context-aware, task-specific schema view that is both compact and semantically rich. It strikes a balance between the exhaustive coverage of full-schema approaches and the minimalism of gold or oracle schemas—offering an effective middle ground that significantly improves SQL generation quality.

Moreover, by preserving only the schema elements grounded in query semantics and validated through two independent reasoning paths, the merged schema enhances robustness and interpretability. It also leads to reduced computational overhead during SQL generation, as the model operates on a tailored schema subset aligned with the user intent.




\section{Experimental Setup}

\subsection{Datasets and Baselines}

\paragraph{Datasets.} We evaluate our schema linking framework on two challenging Text-to-SQL benchmarks with different schema complexity. \textbf{BIRD}~\cite{NEURIPS2023_83fc8fab} includes 95 real-world databases from 37 domains, designed to test execution correctness and implicit reasoning. In this paper, we use the development set for evaluation, which comprises 1,534 text-to-SQL pairs across 11 databases. \textbf{Spider}~\cite{yu-etal-2018-spider} contains 10K questions across 200 databases, emphasizing compositional generalization. The development set contains 1,034 examples spanning 20 complex databases across multiple domains.

\paragraph{Baselines.} We compare our method against a set of strong baselines that reflect the current state of the art in schema-aware and LLM-based Text-to-SQL systems. These include decomposed in-context learning (DIN-SQL~\cite{pourreza2024din}, C3SQL~\cite{dong2023c3zeroshottexttosqlchatgpt}), robust schema linking (RSL-SQL~\cite{cao2024rsl}, TA-SQL~\cite{qu-etal-2024-generation}, Solid-SQL~\cite{liu-etal-2025-solid}), direct linking with question enrichment (E-SQL~\cite{caferoğlu2025esqldirectschemalinking}), and multi-agent or multi-path models such as CHESS~\cite{talaei2024chesscontextualharnessingefficient}, MAC-SQL ~\cite{wang2024macsqlmultiagentcollaborativeframework} and MCS-SQL~\cite{lee-etal-2025-mcs}. Together, these baselines represent diverse and competitive strategies for addressing the challenges of schema linking and SQL generation.

\subsection{Evaluation Strategy}

We evaluate schema retrieval performance using standard metrics: Recall, and False Positive Rate (FPR) \cite{maamari2024the, cao2024rsl, qu-etal-2024-generation}. To obtain the gold schema elements for evaluation, we utilize the \texttt{sqlglot} library~\cite{mao2021sqlglot} to extract the referenced tables and columns from the ground-truth SQL queries.

For SQL generation, we assess performance under three different schema settings. In the \textit{Full Schema} setting, the model is provided with the entire database schema, representing a baseline with no filtering. The \textit{Perfect Schema} setting uses only the gold schema elements parsed from the reference SQL, representing an oracle upper bound. Finally, in the \textit{Retrieved Schema} setting, the model generates SQL queries based on the subset of schema elements selected by our proposed bidirectional schema linking framework. We report SQL Execution Accuracy (EX) as the percentage of generated SQL queries that produce results matching the reference answers. For both the schema linking and SQL generation tasks, we consider only a single response for each LLM call.

\subsection{Implementation Details}

We use four LLMs for evaluation: \texttt{GPT-4o-mini(G)} \cite{openai2024gpt4ocard}, \texttt{DeepSeek-Chat(D)} \cite{deepseekai2025deepseekv3technicalreport}, \texttt{Qwen-2.5-7B-Instruct(Q)} \cite{hui2024qwen25codertechnicalreport}, and \texttt{Gemini-2.0-Flash(M)} \cite{geminiteam2024geminifamilyhighlycapable}. Prompting strategies are adapted from prior work \cite{talaei2024chesscontextualharnessingefficient, maamari2024the}. Schema linking and SQL generation are performed in two stages using structured instructions and augmented query context. Further implementation details and prompt templates are available in Appendix~\ref{appendix:implementation_detail} and \ref{appendix:appendix-prompts}. The implementation of our framework is publicly available at \url{https://github.com/mahadi-nahid/BiSchemaLink}.

\section{Results and Discussion}

In this section, we present a comprehensive evaluation of our proposed bidirectional schema linking framework. Our experiments are designed to assess the effectiveness of schema retrieval and its downstream impact on SQL generation.

\begin{table}[ht]
\centering
\small
\setlength{\tabcolsep}{4pt}
\sisetup{
    detect-all,
    round-mode=places,
    round-precision=2,
    table-align-text-post=false
}
\begin{tabular}{
    l
    | S[table-format=2.2]
    | S[table-format=2.2]
}
\toprule
\textbf{Method} & \textbf{Rec. ($\mathbf{\uparrow}$)} & \textbf{FPR ($\mathbf{\downarrow}$)} \\
\midrule
\rowcolor{red!10}
\textbf{Full Schema}       & 100.00 & 94.62 \\
\midrule
DIN-SQL\cite{pourreza2024din}           & 63.58  & 11.11\\
HySCSL            & 90.36  & 82.08 \\
SCSL              & 88.77  & 67.23 \\
HyTCSL            & 83.00  & 19.84 \\
TCSL              & 77.44  & \textbf{9.79}  \\
TA-SQL \cite{qu-etal-2024-generation}           & 71.90  & \textbf{9.76} \\
RSL-SQL \cite{cao2024rsl}           & \textbf{93.28}  & 68.23 \\
CHESS \cite{talaei2024chesscontextualharnessingefficient}           & \textbf{97.12}  & 30.57 \\

\midrule

\rowcolor{green!10}
\textbf{Ours \texttt{(G)}} & \textbf{90.60}  & \textbf{25.29} \\
\rowcolor{green!10}
\textbf{Ours \texttt{(M)}} & \textbf{92.91}     & \textbf{25.75}   \\
\rowcolor{green!10}
\textbf{Ours \texttt{(D)}} & \textbf{91.56}    & \textbf{19.04}  \\
\rowcolor{green!10}
\textbf{Ours \texttt{(Q)}} & \textbf{85.45}     & \textbf{33.69}  \\
\bottomrule
\end{tabular}
\caption{Schema Linking performance on the BIRD-Dev set. We report Recall (Rec.), and False Positive Rate (FPR) for various schema linking methods. Our approach achieves a strong balance between recall and precision, significantly reducing the false positive rate. Results for other baselines are taken from \cite{maamari2024the, cao2024rsl, qu-etal-2024-generation, talaei2024chesscontextualharnessingefficient}. \texttt{G=GPT-4o-mini}, \texttt{M=Gemini-2.0-flash}, \texttt{D=DeepSeek}, \texttt{Q=Qwen-2.5-7B-Instruct}.}
\label{tab:sl-baselines}
\end{table}

\begin{table*}[ht]
\centering
\small
\setlength{\tabcolsep}{6pt} 
\begin{tabular}{
    >{\raggedright\arraybackslash}p{2.9cm} 
    | >{\raggedright\arraybackslash}p{3cm} 
    | S[table-format=2.2]
    | S[table-format=2.2]
    | S[table-format=2.2]
    | S[table-format=2.2]
    | S[table-format=2.2]} 
\toprule
\textbf{LLM} & \textbf{Method} & \textbf{Simple} & \textbf{Moderate} & \textbf{Challenging} & \textbf{Total} & \textbf{$\mathbf{\Delta}$ ($\mathbf{\uparrow}$$\mathbf{\downarrow}$)} \\
\midrule

\multirow{3}{*}{\textbf{\texttt{GPT-4o-mini(G)}}} 
& Full Schema     & {54.27} & {34.56} & {26.90} & {45.70} & \textemdash \\
& Perfect Schema  & {67.46} & {42.55} & {35.17} & {56.84} & {+11.14 }\\
\rowcolor{gray!10}
& \textbf{Retrieved Schema} &  {58.70 } &  {36.72} &  {33.79} & \textbf{51.11} &  \textbf{+5.41 } $\mathbf{\uparrow}$ \\
\midrule
 
\multirow{3}{*}{\textbf{\texttt{Gemini-2.0-flash(M)}}} 
& Full Schema     & {59.03} & {42.67} & {37.93} & {52.09} & \textemdash \\
& Perfect Schema  & {68.65} & {49.57} & {45.52} & {60.69} & {+8.60} \\
\rowcolor{gray!10}
& \textbf{Retrieved Schema} & {63.46} & {44.92} & {43.45} & \textbf{57.86} & \textbf{+5.77 $\mathbf{\uparrow}$} \\
\midrule

\multirow{3}{*}{\textbf{\texttt{DeepSeek(D)}}}         
& Full Schema     & {61.43} & {38.61} & {41.18 } & {53.80} & \textemdash \\   
& Perfect Schema  & {71.46} & {52.59} & {51.03 } & {63.82} & {+10.02} \\
\rowcolor{gray!10}                                 
& \textbf{Retrieved Schema} & { 62.35} & {45.17} & {49.41} & \textbf{56.80} & \textbf{+3.00  $\mathbf{\uparrow}$}\\
\midrule

\multirow{3}{*}{\textbf{\texttt{Qwen-2.5-7B(Q)}}} 
& Full Schema     & {43.14 } & {27.59} & {25.52} & {37.13} & \textemdash \\
& Perfect Schema  & {56.86} & {35.78} & {35.17} & {48.44 } & {+11.31} \\
\rowcolor{gray!10}
& \textbf{Retrieved Schema} & {47.35} & {29.96} & {26.90} & \textbf{40.16} & \textbf{ +3.03 $\mathbf{\uparrow}$} \\


\bottomrule
\end{tabular}
\caption{Zero-shot execution accuracy on the BIRD-Dev set across different SQL hardness levels: Simple, Moderate, and Challenging. We compare performance using the full schema, perfect (oracle) schema, and our retrieved schema. The "$\mathbf{\Delta}$ ($\mathbf{\uparrow}$$\mathbf{\downarrow}$)" column reports improvement in total execution accuracy over the Full Schema baseline.}
\label{tab:sql-ex-performance-bird}
\end{table*}
\subsection{Schema Linking Performance}

Table~\ref{tab:sl-baselines} presents the schema linking performance of various approaches on the BIRD development set, with a focus on two critical metrics: Recall (Rec.) and False Positive Rate (FPR). In Text-to-SQL systems, achieving high recall ensures that all relevant schema elements are available for query generation, while a low FPR is essential to prevent spurious tokens that can mislead large language models (LLMs). Our proposed method consistently achieves an optimal balance between these two objectives across all LLM variants compared to Table-then-Column Schema Linking (TCSL), Single Column Schema Linking (SCSL), Hybrid TCSL and Hybrid SCSL \cite{talaei2024chesscontextualharnessingefficient, maamari2024the}.

A key observation from the results is the recurring trade-off in prior work between recall and FPR. For instance, CHESS~\cite{talaei2024chesscontextualharnessingefficient} achieves the highest recall (97.12\%) but at the cost of an elevated false positive rate (30.57\%). Moreover, CHESS requires an average of 78 LLM calls and processes approximately 339,965 tokens per request on the BIRD dev set~\cite{chung2025longcontextneedleveraging}, posing significant challenges for inference efficiency and scalability. Similarly, RSL-SQL~\cite{cao2024rsl} reports strong recall (93.28\%) yet suffers from a high FPR (68.23\%), indicating over-inclusion of irrelevant schema components. Such excessive linking inflates the prompt size, introduces unnecessary context, and ultimately degrades SQL generation accuracy by increasing the likelihood of query hallucinations. 

In contrast, our bidirectional schema linking framework demonstrates that it is possible to achieve competitive recall, for instance {92.91\%} with Gemini (M) and {91.56\%} with DeepSeek (D) while drastically reducing false positive rates to {19.04\%} and {25.75\%}, respectively. This reduction is not only statistically but also practically beneficial. A lower FPR directly translates to fewer irrelevant columns and tables in the prompt, leading to improved interpretability, lower token usage, and more accurate SQL generation.

We present additional results on the \textbf{Spider} datasets and detailed error analysis in \textbf{Appendix~\ref{appendix:additional_results}} and \textbf{Appendix~\ref{appendix:error_analysis}}, further demonstrating the effectiveness of our approach across diverse schema structures.

\subsection{Impact on Text-to-SQL Performance}

Table~\ref{tab:sql-ex-performance-bird} reports zero-shot execution accuracy across different schema access settings—Full Schema, Perfect Schema, and our Retrieved Schema—on the BIRD development set, categorized by SQL hardness levels (Simple, Moderate, and Challenging). Across all evaluated LLMs, our \textbf{retrieved schema} consistently outperforms the full schema baseline. For instance, with \texttt{GPT-4o-mini} and \texttt{Gemini-2.0-flash}, our method achieves total execution accuracies of {51.11\%} and {57.86\%}, respectively, effectively reducing the performance gap between the full schema baseline and the oracle (perfect schema) by approximately \textbf{50\%}. Notably, these substantial improvements of \textbf{5.41\%} and \textbf{5.77\%} over the full schema are achieved without employing any SQL query fixing or correction strategies. Similarly, \texttt{DeepSeek} and \texttt{Qwen} exhibit gains of {3.00\%} and {3.03\%}, respectively. These results provide strong empirical evidence that filtering out irrelevant tables and columns via our bidirectional schema linking mechanism reduces prompt noise and enhances the model’s ability to generate correct SQL queries. 

In our experiments, \texttt{Qwen} exhibited relatively lower performance compared to other models. As a smaller model, it may have limited generalization and reasoning capacity on complex benchmarks.  Additionally, Qwen’s schema linking capabilities were not as strong as those of the other models, further contributing to its lower results. Smaller models such as \texttt{Qwen-2.5-7B-Instruct} tend to be more sensitive to prompt design and often require more careful engineering to achieve optimal results.


\paragraph{Comparison with other Models.}

Table~\ref{tab:sql-ex-baselines-bird} presents a comparison of zero-shot SQL execution accuracy on the BIRD development set between our approach and several baseline methods. Our approach achieves performance on par with strong baselines such as E-SQL and Solid-SQL. However, it is important to highlight a critical difference: unlike these methods, our framework does not incorporate any SQL query fixing, refinement, or self-consistency techniques during SQL generation. In contrast, most existing methods depend on post-generation correction pipelines to boost execution accuracy~\cite{talaei2024chesscontextualharnessingefficient, pourreza2024din, gao2024xiyan, lee-etal-2025-mcs, caferoğlu2025esqldirectschemalinking, wang2024macsqlmultiagentcollaborativeframework}.


\begin{table}[ht]
\centering
\setlength{\tabcolsep}{8pt}
\renewcommand{\arraystretch}{1.1}
\small
\begin{tabular}{
    >{\raggedright\arraybackslash}p{2cm}
    | S[table-format=2.1]
    | S[table-format=2.1]}
\toprule
\textbf{Method} & \textbf{Model} & \textbf{EX (\%$\mathbf{\uparrow}$)} \\
\midrule
GPT-4      & \texttt{GPT-4}  & 46.2 \\
{C3SQL *}      & \texttt{GPT-3.5}  & 50.2 \\
{DIN-SQL *}    & \texttt{GPT-4}  & 50.6 \\
{DAIL-SQL *}   & \texttt{GPT-4}   & 54.8 \\
{o1-mini}   & \texttt{o1-mini}   & 52.4 \\
{o3-mini}   & \texttt{o3-mini}   & 53.7 \\
\midrule
{Solid-SQL *}   & \texttt{GPT-4o-mini}   & 58.9 \\
{E-SQL *}       & \texttt{GPT-4o-mini}   & 56.6 \\
RSL-SQL    & \texttt{GPT-4o}   & 48.7 \\
RSL-SQL    & \texttt{DeepSeek}   & 42.9 \\
{MAC-SQL *}     & \texttt{GPT-4o-mini}   & 50.6 \\
{MCS-SQL *}     & \texttt{GPT-4}          & 50.7 \\ 
{TA-SQL *}      & \texttt{GPT-4-turbo}    & {52.4} \\
\midrule
\rowcolor{teal!10}
\textbf{Ours \texttt{(G)}}   & \texttt{GPT-4o-mini}    & \textbf{51.1} \\
\rowcolor{teal!10}
\textbf{Ours \texttt{(M)}}   & \texttt{Gemini-2.0-flash}    & \textbf{57.8} \\
\rowcolor{teal!10}
\textbf{Ours \texttt{(D)}}    & \texttt{DeepSeek-chat} & \textbf{56.8} \\
\rowcolor{teal!10}
\textbf{Ours \texttt{(Q)}}    & \texttt{Qwen-2.5-7B} & \textbf{40.1} \\
\bottomrule
\end{tabular}
\caption{Zero-shot execution accuracy on the BIRD development set compared to various baselines. Notably, some baselines employ query refinement, fixing, or self-consistency techniques in their zero-shot experiments (*). In contrast, our approach evaluates a single response from the LLM without utilizing multi-response selection, self-correction, or query fixing strategies.}

\label{tab:sql-ex-baselines-bird}
\end{table}
\vspace{-0.5em}

\paragraph{Plug-in with other Models.}

To further assess the utility and portability of our retrieved schema, we conducted plug-in experiments on the BIRD dataset using two representative Text-to-SQL models: DIN-SQL \cite{pourreza2024din} and MCS-SQL \cite{lee-etal-2025-mcs}. The results is summarized in Table \ref{tab:plugin-results}. In these experiments, we replaced each model’s default schema linking component with our retrieved schema, keeping all other components unchanged. The results demonstrate notable improvements in execution accuracy. DIN-SQL improved from 50.6\% to 55.4\%, while MCS-SQL improved from 52.8\% to 54.7\%. These gains confirm the effectiveness of our method as a drop-in schema retrieval module, enhancing downstream performance across different architectures. 

\begin{table}[h]
\centering
\small
\begin{tabular}{l|c|c}
\toprule
\textbf{Model} & \textbf{Acc (\%)} & \textbf{W/ Our Schema (\%)} \\
\midrule
DIN-SQL & 50.6 & 55.4 \\
MCS-SQL & 52.8 & 54.7 \\
\bottomrule
\end{tabular}
\caption{Plug-in results on the BIRD dataset: Execution accuracy with and without our retrieved schema.}
\label{tab:plugin-results}
\end{table}


\subsection{Ablation Study}

To better understand the contributions of individual components within our bidirectional schema linking framework, we conduct an ablation study across four LLMs (GPT-4o-mini, Gemini, DeepSeek, and Qwen), focusing on Schema Linking Recall (SLR) and False Positive Rate (FPR), the two most critical factors for effective and efficient Text-to-SQL generation. As shown in Table~\ref{tab:Ablation}, the full \textbf{bidirectional} approach consistently delivers the best trade-off across models, achieving the highest recall ({92.91\%} for Gemini and {91.56\%} for DeepSeek), while maintaining relatively low FPRs ({25.75\%} and {19.04\%}, respectively). This demonstrates the robustness and generalizability of our framework across diverse LLM backbones.

\begin{table}[ht]
\centering
\small
\setlength{\tabcolsep}{4pt}
\sisetup{
    detect-all,
    round-mode=places,
    round-precision=2,
    table-align-text-post=false
}
\begin{tabular}{
    l
    | S[table-format=2.2]
    | S[table-format=2.2]
}
\toprule
\textbf{Method} & \textbf{Rec. ($\mathbf{\uparrow}$)} & \textbf{FPR ($\mathbf{\downarrow}$)} \\

\midrule

\rowcolor{gray!10}
\textbf{Bidirectional \texttt{(G)}} & \textbf{90.60}  & \textbf{25.29}  \\

\rowcolor{gray!5}
{w/o Augmentation \texttt{(G)}} & {87.70}    & {28.80} \\

\rowcolor{gray!5}
{w/o Table First \texttt{(G)}} & {88.15}    & {23.33}   \\

\rowcolor{gray!5}
{w/o Column First \texttt{(G)}} & {81.65}     & {18.81}   \\

\midrule

\rowcolor{gray!10}
\textbf{Bidirectional \texttt{(M)}} & \textbf{92.91}     & \textbf{25.75}   \\

\rowcolor{gray!5}
{w/o Augmentation \texttt{(M)}}  & {89.76}    & {22.39}   \\

\rowcolor{gray!5}
{w/o Table First \texttt{(M)}}  & {91.13}    & {19.94}  \\

\rowcolor{gray!5}
{w/o Column First \texttt{(M)}} & {87.90}    & {17.93}   \\

\rowcolor{gray!5}
\midrule

\rowcolor{gray!10}
\textbf{Bidirectional \texttt{(D)}} & \textbf{91.56}    & \textbf{19.04}   \\

\rowcolor{gray!5}
{w/o Augmentation \texttt{(D)}}  & {88.03}    & {20.37}   \\
					
\rowcolor{gray!5}
{w/o Table First \texttt{(D)}} & {88.06}     & {14.16}   \\

\rowcolor{gray!5}
{w/o Column First \texttt{(D)}} & {89.49}     & {18.49}  \\
\midrule

\rowcolor{gray!10}
\textbf{Bidirectional \texttt{(Q)}} & \textbf{85.45}     & \textbf{33.69}   \\

\rowcolor{gray!5}
{w/o Augmentation \texttt{(Q)}}  & {83.78}    & {31.46}   \\
				
\rowcolor{gray!5}
{w/o Table First \texttt{(Q)}} & {79.54}    & {26.38}   \\

\rowcolor{gray!5}
{w/o Column First \texttt{(Q)}}& {66.97}    & {22.31}   \\

\bottomrule

\end{tabular}
\caption{Ablation study evaluating different schema linking strategies in terms of Schema Linking Recall (SLR) and False Positive Rate (FPR). The Bidirectional approach, which combines both Table-First and Column-First outputs, achieves the best overall performance.} 
\label{tab:Ablation}
\end{table}


We observe that removing either the Table-First or Column-First module leads to noticeable recall degradation. For instance, with GPT-4o-mini, eliminating the Column-First step drops recall from {90.60\%} to {81.65\%} a substantial decrease of nearly 9 percentage points while slightly reducing FPR (from 25.29\% to 18.81\%). This suggests that although a simpler schema linking module might appear more precise, it risks missing important schema elements, ultimately harming SQL generation coverage. Conversely, removing the Table-First step results in a smaller drop in recall (from {92.91\%} to {91.13\%} for Gemini), but again leads to a reduction in FPR. These findings reveal that the Table-First and Column-First modules offer complementary strengths: the former enhances schema-level connectivity, while the latter fine-tunes column-level relevance.


\begin{table}[ht]
\centering
\small
\setlength{\tabcolsep}{10pt}
\sisetup{
    detect-all,
    round-mode=places,
    round-precision=2,
    table-align-text-post=false
}
\begin{tabular}{
    l
    | S[table-format=2.2]
    | S[table-format=2.2]
}
\toprule
\textbf{Method} & \textbf{Tab (\%$\mathbf{\downarrow}$)} & \textbf{Col (\%$\mathbf{\downarrow}$)}\\
\midrule
\rowcolor{green!10}
\textbf{Perfect Schema} & {67.60}    & {89.39} \\  
\midrule

\rowcolor{gray!10}
\textbf{Bidirectional \texttt{(G)}} & {63.45}    & {84.29}  \\

\rowcolor{gray!5}
{w/o Table First \texttt{(G)}}  & {64.45}    & {85.99}  \\

\rowcolor{gray!5}
{w/o Column First \texttt{(G)}} & {65.66}    & {86.62} \\

\midrule

\rowcolor{gray!10}
\textbf{Bidirectional \texttt{(M)}} & {63.41}    & {83.49} \\

\rowcolor{gray!5}
{w/o Table First \texttt{(M)}}  & {64.45}    & {85.99} \\

\rowcolor{gray!5}
{w/o Column First \texttt{(M)}} & {65.66}    & {86.62}  \\

\rowcolor{gray!5}
\midrule

\rowcolor{gray!10}
\textbf{Bidirectional \texttt{(D)}} & {63.70}    & {86.57}  \\

\rowcolor{gray!5}
{w/o Table First \texttt{(D)}} & {64.34}    & {86.99} \\

\rowcolor{gray!5}
{w/o Column First \texttt{(D)}} & {65.57}    & {88.38} \\
\midrule

\rowcolor{gray!10}
\textbf{Bidirectional \texttt{(Q)}} & {60.08}    & {81.51} \\

\rowcolor{gray!5}
{w/o Table First \texttt{(Q)}} & {64.44}    & {79.82} \\

\rowcolor{gray!5}
{w/o Column First \texttt{(Q)}} & {66.36}    & {87.63} \\

\bottomrule

\end{tabular}
\caption{Schema reduction performance for different method variants. We report the percentage of tables and columns retained after applying schema linking using each approach. Lower percentages indicate greater reduction, contributing to more efficient and focused Text-to-SQL generation.}
\label{tab:schema_reduction}
\end{table}

\subsection{Schema Reduction and Error Analysis}
We evaluate the effectiveness of our proposed methods in reducing the database schema size by measuring the percentage of tables and columns retained after applying schema linking. Table~\ref{tab:schema_reduction} reports the reduction rates for different variants of our method, including the full bidirectional approach, table-first, and column-first approach. The results demonstrate that our bidirectional schema linking strategy is effective at significantly narrowing down the schema context by selecting only the most relevant tables and columns. This targeted reduction not only simplifies the reasoning space for the language model but also improves execution efficiency, especially for databases with large schemas. These findings highlight the practical advantage of incorporating schema pruning into the Text-to-SQL pipeline.


Illustrative examples are presented in \textbf{Appendix~\ref{appendix:illustrative_example_cases}}, detailed error cases are discussed in \textbf{Appendix~\ref{appendix:error_example_case}}, and a comprehensive error analysis is provided in \textbf{Appendix~\ref{appendix:error_analysis}}. 
The appendices systematically categorize common failure modes and include representative examples from our experiments. 

\subsection{Token-LLM-calls-latency Analysis}


\begin{table*}[ht]
\centering
\small
\setlength{\tabcolsep}{8pt}
\begin{tabular}{l|c|c|c}
\toprule
\textbf{Method} & \textbf{\# LLM Calls} & \textbf{Tokens / Sample} & \textbf{Latency (s)} \\
\midrule
CHESS~\cite{talaei2024chesscontextualharnessingefficient} & 78 & 339k & 46.8 \\
\midrule
DIN-SQL \cite{pourreza2024din} & 1 & 8.7k & 1.2 \\
\midrule
SCSL~\cite{maamari2024the} & 78 & 339k & 46.8 \\
\midrule
HyTCSL~\cite{maamari2024the} & 78 & 340k & 46.8 \\
\midrule
TCSL~\cite{maamari2024the} & 2 & 8.7k & 1.6 \\
\midrule
RSL-SQL \cite{cao2024rsl} & 2 & 8.7k & 1.6 \\
\midrule
TA-SQL\cite{qu-etal-2024-generation} & 1 & 6.1k & 0.75 \\
\midrule
\textbf{Ours} & 6 & 10.5k & 2.6 \\
\bottomrule
\end{tabular}
\caption{Comparison of inference cost (\# llm calls, tokens, latency) on the BIRD dataset.}
\label{tab:inference-cost}
\end{table*}


we conducted a detailed empirical analysis comparing LLM inference costs across several competitive baselines using the BIRD dataset. Specifically, we measured the number of LLM calls, total input tokens per sample, and inference latency using the GPT-4o-mini tokenizer for consistency. As shown in Table \ref{tab:inference-cost}, our method significantly reduces token consumption and latency compared to multi-stage methods such as CHESS, SCSL, and HyTCSL, which require approximately 78 LLM calls and over 300k tokens per query, resulting in latency close to 47 seconds. In contrast, our framework achieves a balanced trade-off with just 6 LLM calls, 10.5k tokens per sample, and a latency of 2.6 seconds—demonstrating a practical compromise between efficiency and performance. We believe this analysis strengthens the case for the scalability and deployability of our system in real-world applications.

\subsection{Discussion}
Our bidirectional schema linking approach effectively addresses schema selection challenges in Text-to-SQL by leveraging LLMs through a structured combination of table-first and column-first retrieval strategies. This dual strategy refines schema selection, reducing irrelevant elements while preserving high recall. Experiments show that it maintains competitive accuracy with significantly fewer false positives, improving SQL generation by minimizing schema-induced distractions—especially crucial for large-scale databases.

A major strength of our method lies in its simplicity and generalizability. Unlike methods requiring fine-tuning or costly similarity computations~\cite{liu2024solidsqlenhancedschemalinkingbased}, our approach is fully prompt-based and easily adaptable across LLMs. Its multi-stage retrieval efficiently reduces schema size and LLM calls, enhancing practical deployment. Notably, it may boost the performance of smaller LLMs by simplifying schema context, enabling accurate SQL generation even in constrained environments. Our results demonstrate that effective schema linking is crucial for improving Text-to-SQL generation. By leveraging our retrieved schema, we significantly enhance the LLM’s SQL query generation performance—all without relying on multiple selection, self-consistency, or the complex query refinement strategies prevalent in recent work. An interesting direction for future work is to develop adaptive retrieval strategies that estimate query complexity and selectively execute only a single retrieval path in simpler scenarios, aiming to reduce redundant LLM calls while maintaining accuracy. 



\section{Conclusion}
We propose a novel bidirectional schema retrieval framework that re-frames schema linking as a standalone task within the Text-to-SQL pipeline. By combining table-first and column-first retrieval strategies with augmentation techniques such as question decomposition and keyword extraction, our approach significantly improves schema linking precision and recall while reducing false positives. Experiments on the BIRD and Spider dataset demonstrate notable gains in SQL generation accuracy and computational efficiency. By treating schema linking as an independent task, our work provides a generalizable and reusable framework for retrieval-based NLP applications. Future work can explore adaptive retrieval strategies for further refinement.

\section*{Acknowledgements}
We sincerely thank the reviewers for their thoughtful feedback, insightful suggestions, and encouraging remarks, which helped improve the quality of this work. This research was supported in part by the Natural Sciences and Engineering Research Council of Canada (NSERC). Md Mahadi Hasan Nahid was supported by the Alberta Innovates Graduate Student Scholarship. This work was conducted during an internship at Huawei Technologies Canada Co., Ltd., Burnaby, Canada.

\section*{Limitations}

Despite the effectiveness of our bidirectional schema linking approach in improving schema retrieval and SQL generation, certain limitations remain.

First, our method relies on large language models (LLMs) for both schema linking and question augmentation. While LLMs demonstrate strong reasoning capabilities, their responses are non-deterministic and may introduce inconsistencies in schema selection, especially for complex or ambiguous queries.

Second, our retrieval framework requires multiple LLM calls, which can increase latency and computational costs compared to single-pass retrieval methods. Although our approach significantly reduces token usage compared to full schema settings, further optimization is needed to minimize inference time.

Finally, while our bidirectional schema retrieval improves schema linking recall while controlling false positives, it does not guarantee optimal selection in all cases. Certain database schemas with highly similar column names across multiple tables may still lead to selection errors, affecting SQL generation accuracy. Addressing these limitations in future research could further enhance schema retrieval efficiency and improve LLM-driven SQL generation in complex database environments.

\section*{Ethical Considerations} 
The reliance on large language models (LLMs) for schema retrieval and SQL generation in our approach necessitates careful consideration of ethical implications. First, LLMs may inherit biases from training data, potentially leading to unfair or inaccurate schema selection. Second, while our study does not use sensitive data, real-world applications must ensure privacy and security when handling proprietary database schemas. Third, the computational cost of multiple LLM calls contributes to energy consumption, necessitating more efficient retrieval strategies. Finally, although we evaluate on public datasets, real-world deployments require careful consideration of fairness and unintended information exposure. Addressing these concerns is essential for responsible AI development in database-related tasks.

\bibliography{custom}

\clearpage
\appendix
\twocolumn[{%
 \centering
 \Large\bf Supplementary Material: Appendices \\ [20pt]
}]

\section{Additional Results}
\label{appendix:additional_results}

\paragraph{Results on Spider}
We perform our schema linking approach on Spider \cite{yu-etal-2018-spider} dev dataset.
Our bidirectional schema linking approach consistently achieves high recall and strong F1 scores across all evaluated models on the Spider dataset (Table~\ref{tab:performacne_spider-1}). This demonstrates the method’s effectiveness in retrieving relevant schema elements while maintaining a favorable balance between recall and precision. The results confirm that leveraging both table-first and column-first retrieval provides robust schema linking performance.
\begin{table}[ht]
\centering
\small
\setlength{\tabcolsep}{4pt}
\sisetup{
    detect-all,
    round-mode=places,
    round-precision=2,
    table-align-text-post=false
}
\begin{tabular}{
    l
    | S[table-format=2.2]
    | S[table-format=2.2]
}
\toprule
\textbf{Method} & \textbf{Rec. ($\mathbf{\uparrow}$)} & \textbf{FPR ($\mathbf{\downarrow}$)} \\

\midrule

\rowcolor{gray!10}
\textbf{Bidirectional \texttt{(G)}} & \textbf{92.35}	& \textbf{33.3} \\
\rowcolor{gray!5}
{w/o Table First \texttt{(G)}} & {91.05}	& {31.35}	\\
\rowcolor{gray!5}
{w/o Column First \texttt{(G)}} & {89.29}	& {25.06}	\\
\midrule


\rowcolor{gray!10}
\textbf{Bidirectional \texttt{(M)}} & \textbf{92.55}    & \textbf{32.36}   \\
\rowcolor{gray!5}
{w/o Table First \texttt{(M)}} & {91.61}    & {29.76}    \\
\rowcolor{gray!5}
{w/o Column First \texttt{(M)}} & {89.23}    & {24.1}   \\
\midrule


\rowcolor{gray!10}
\textbf{Bidirectional \texttt{(D)}} & \textbf{89.35}    & \textbf{24.32}   \\
\rowcolor{gray!5}
{w/o Table First \texttt{(D)}} & {88.95}    & {22.04}    \\
\rowcolor{gray!5}
{w/o Column First \texttt{(D)}} & {88.39}    & {17.11}   \\
\midrule


\rowcolor{gray!10}
\textbf{Bidirectional \texttt{(Q)}} & \textbf{86.19}    & \textbf{34.58}  \\
\rowcolor{gray!5}
{w/o Table First \texttt{(Q)}} & {82.26}    & {30.54}  \\
\rowcolor{gray!5}
{w/o Column First \texttt{(Q)}} & {80.33}    & {22.26}   \\

\bottomrule
\end{tabular}
\caption{Schema Linking Recall (SLR) and False Positive Rate (FPR) on Spider Dataset. Lower FPR and higher SLR indicate better schema retrieval effectiveness.}
\label{tab:performacne_spider-1}
\end{table}

In the ablation studies, removing either the table-first or column-first component leads to changes in retrieval dynamics. The bidirectional design thus provides complementary benefits: Table-First improves coverage of relevant entities, while Column-First fine-tunes precision by focusing on attribute-level cues. While the results indicate that the impact of the Table-First and Column-First strategies is modest, this may be attributed to the nature of the Spider dataset, which is less complex than BIRD. 


\begin{table}[ht]
\centering
\small
\setlength{\tabcolsep}{8pt}
\sisetup{
    detect-all,
    round-mode=places,
    round-precision=2,
    table-align-text-post=false
}
\begin{tabular}{
    l
    | S[table-format=2.2]
    | S[table-format=2.2]
}
\toprule
\textbf{Method} & \textbf{Tab (\%$\mathbf{\downarrow}$)} & \textbf{Col (\%$\mathbf{\downarrow}$)} \\
\midrule
\rowcolor{green!10}
\textbf{Perfect Schema} & {54.57}    & {79.93}   \\ 
\midrule

\rowcolor{gray!10}
\textbf{Bidirectional \texttt{(G)}} & {51.85}    & {73.48}  \\ 

\rowcolor{gray!5}
{w/o Table First \texttt{(G)}}  & {51.78}    & {74.74}  \\ 

\rowcolor{gray!5}
{w/o Column First \texttt{(G)}} & {52.79}    & {78.81} \\ 

\midrule

\rowcolor{gray!10}
\textbf{Bidirectional \texttt{(M)}} & {50.93}    & {73.46} \\ 

\rowcolor{gray!5}
{w/o Table First \texttt{(M)}}  & {51.51}    & {75.56}  \\ 

\rowcolor{gray!5}
{w/o Column First \texttt{(M)}} & {55.24}    & {77.40}   \\ 

\rowcolor{gray!5}
\midrule

\rowcolor{gray!10}
\textbf{Bidirectional \texttt{(D)}} & {50.35}    & {71.88} \\ 

\rowcolor{gray!5}
{w/o Table First \texttt{(D)}} & {50.8}    & {72.63}  \\ 

\rowcolor{gray!5}
{w/o Column First \texttt{(D)}} & {53.03}    & {75.70} \\ 
\midrule

\rowcolor{gray!10}
\textbf{Bidirectional \texttt{(Q)}} & {50.53}    & {72.78}  \\ 

\rowcolor{gray!5}
{w/o Table First \texttt{(Q)}} & {53.23}    & {81.02} \\ 

\rowcolor{gray!5}
{w/o Column First \texttt{(Q)}} & {53.22}    & {75.85} \\ 

\bottomrule

\end{tabular}
\caption{Schema reduction performance for different method variants on Spider. We report the percentage of tables and columns retained after applying schema linking. Lower percentages indicate more effective schema pruning, contributing to efficient SQL generation.}

\label{tab:schema_reduction_spider-1}
\end{table}


\begin{table*}[ht]
\centering
\small
\setlength{\tabcolsep}{10pt} 
\begin{tabular}{
    >{\raggedright\arraybackslash}p{2.8cm} 
    | >{\raggedright\arraybackslash}p{3cm} 
    | S[table-format=2.2]
    | S[table-format=2.2]
    | S[table-format=2.2]
    | S[table-format=2.2]
    | S[table-format=2.2]} 
\toprule
\textbf{LLM} & \textbf{Method} & \textbf{Easy} & \textbf{Medium} & \textbf{Hard} & \textbf{Extra} & \textbf{All $\mathbf{\Delta}$ ($\mathbf{\uparrow}$$\mathbf{\downarrow}$)} \\
\midrule

\multirow{3}{*}{\textbf{\texttt{GPT-4o-mini(G)}}} 
& Full Schema     & {85.0} & {70.6} & {50.0} & {38.0} & {64.5}\\
& Perfect Schema  & {88.3} & {70.9} &{59.2} & {38.0} & {66.9 ($\mathbf{\uparrow}$ 2.4)} \\
 
\rowcolor{gray!10}
& \textbf{Retrieved Schema} & {89.3} & {71.1} &{54.6} & {33.7} & \textbf{65.7 ($\mathbf{\uparrow}$ 1.2)} \\
\midrule

\multirow{3}{*}{\textbf{\texttt{Gemini-2.0-flash(M)}}} 
& Full Schema     & {82.0} & {77.4} & {55.7} & {41.2} & {67.7}\\
& Perfect Schema  & {89.8} & {76.0 } & {66.2} & {44.6} & {74.6 ($\mathbf{\uparrow}$ 6.9)}\\

\rowcolor{gray!10}
& \textbf{Retrieved Schema} &  {88.8} &  {74.4} &  {64.4} & {44.0} &  \textbf{70.6 ($\mathbf{\uparrow}$ 2.9)} \\
\midrule

\multirow{3}{*}{\textbf{\texttt{DeepSeek(D)}}}         
& Full Schema    & {88.2} & {72.5} & {54.0} & {35.5} & {66.3}\\
& Perfect Schema   & {86.8} & {75.2} & {56.3} & {45.2} & {69.2 ($\mathbf{\uparrow}$ 2.9)} \\
\rowcolor{gray!10}                                 
& \textbf{Retrieved Schema}  & {88.7} & {72.3} & {55.7} & {41.6} & \textbf{67.6 ($\mathbf{\uparrow}$ 1.3)} \\


\bottomrule

\end{tabular}
\caption{Zero-shot execution accuracy on the Spider-Dev set across different SQL hardness levels: Easy, Medium, Hard and Extra Hard. We compare performance using the full schema, perfect (oracle) schema, and our retrieved schema. The " $\mathbf{\Delta}$ ($\mathbf{\uparrow}$$\mathbf{\downarrow}$)" column reports improvement in total execution accuracy over the Full Schema baseline.}
\label{tab:sql-ex-performance}
\end{table*}

Table~\ref{tab:schema_reduction_spider-1} shows the schema reduction results in terms of table and column retention rates. Compared to the perfect schema, our bidirectional approach achieves comparable or even better pruning effectiveness. For instance, with DeepSeek, our method retains only {50.35\%} of the tables and {71.88\%} of the columns, closely aligning with the perfect schema's retention rates of {54.57\%} (tables) and {79.93\%} (columns). This indicates that our method not only retrieves relevant schema components effectively but also avoids excessive context that could overwhelm the LLM.

Table~\ref{tab:sql-ex-performance} summarizes zero-shot execution accuracy on the Spider development set across varying SQL hardness levels. Our retrieved schema method consistently outperforms the full schema baseline for all evaluated LLMs, achieving notable gains in overall accuracy. For example, with GPT-4o-mini, our approach increases total execution accuracy by 1.2 points over the full schema, while Gemini-2.0-flash and DeepSeek see improvements of 2.9 and 1.3 points, respectively. Notably, these improvements are achieved without any post-generation query correction, demonstrating the effectiveness and efficiency of our schema retrieval strategy. 

While the effectiveness of our schema linking framework on other complex benchmark, we observe limited improvements on the Spider dataset. This limited gain can be attributed to the nature of Spider, where many queries already focus on well-defined, simpler database contexts, reducing the potential benefits of schema filtering. 


\begin{table}[ht]
\centering
\setlength{\tabcolsep}{1pt}
\renewcommand{\arraystretch}{1.1}
\small
\begin{tabular}{
    >{\raggedright\arraybackslash}p{2.5cm}
    | S[table-format=2.1]
    | S[table-format=2.1]}
\toprule
\textbf{Method} & \textbf{Model} & \textbf{EX (\%$\mathbf{\uparrow}$)} \\
\midrule
GPT-4 \cite{pourreza2024din}      &\texttt{GPT-4}  & {64.9} \\ \midrule
ChatGPT      & \texttt{GPT-3.5-turbo}  & {60.1}\\
\midrule
\rowcolor{blue!10}
\textbf{Ours\texttt{(G)}}   & \texttt{GPT-4o-mini}    & \textbf{65.7} \\
\rowcolor{blue!10}
\textbf{Ours\texttt{(M)}}   & \texttt{Gemini-2.0-flash}    & \textbf{70.6} \\
\rowcolor{blue!10}
\textbf{Ours\texttt{(D)}}    &\texttt{DeepSeek-chat} & \textbf{67.6} \\
\bottomrule
\end{tabular}
\caption{Zero-shot execution accuracy on the Spider development set compared to various baselines.}

\label{tab:sql-ex-baselines-spider}
\end{table}

\paragraph{Effect of Sample Column Values}

In our study, we explored the impact of different schema representations on Schema Linking performance, specifically focusing on the use of sample rows, sample column values, and distinct sample column values. We found that distinct sample column values significantly boosted performance compared to using sample rows or sample column values. This is because when using sample rows, the linear representation of rows makes it difficult for the large language model (LLM) to correctly identify column values. On the other hand, while sample column values can be useful, they may also include redundant values that do not provide additional meaningful information for the LLM. Our results demonstrate that distinct sample values help reduce redundancy and improve schema linking accuracy, leading to a more effective schema retrieval process.

\begin{table}[h]
\centering
\setlength{\tabcolsep}{2pt}
\small
\begin{tabular}{
    >{\raggedright\arraybackslash}p{4.5cm}
    | S[table-format=2.2]
    | S[table-format=2.2]
}
\toprule
\textbf{Column Value Selection Method} & \textbf{SLR (\%$\mathbf{\uparrow}$)} & \textbf{FPR (\%$\mathbf{\downarrow}$)} \\
\midrule
Sample Rows                  & {82.34} & {30.12} \\
\midrule
Sample Column Values         & {86.67} & {22.89} \\
\midrule
\textbf{Distinct Sample Column Values} & \textbf{90.60} & \textbf{25.29} \\
\bottomrule
\end{tabular}
\caption{Schema linking performance using different column value sampling strategies on BIRD dataset. SLR denotes schema linking recall, and FPR represents false positive rate.}
\label{tab:schema_linking_sampling}
\end{table}

\begin{table}[ht]
\centering
\small
\setlength{\tabcolsep}{8pt}
\sisetup{
    detect-all,
    round-mode=places,
    round-precision=2,
    table-align-text-post=false
}
\begin{tabular}{
    l
    | S[table-format=2.2]
    | S[table-format=2.2]
}
\toprule
\textbf{Method} & \textbf{BIRD} & \textbf{Spider}  \\
\midrule

\rowcolor{gray!20}
\textbf{Bidirectional \texttt{(G)}} & \textbf{97.56}  & \textbf{95.29}\\
\rowcolor{gray!5}
{w/o Table First \texttt{(G)}} & {96.62}     & {95.29}   \\
\rowcolor{gray!5}
{w/o Column First \texttt{(G)}} & {96.95}    & {95.20}    \\
\midrule

\rowcolor{gray!20}
\textbf{Bidirectional \texttt{(M)}} & \textbf{97.22}     & \textbf{95.87}    \\
\rowcolor{gray!5}
{w/o Table First \texttt{(M)}}   & {96.37}     & {95.87}     \\
\rowcolor{gray!5}
{w/o Column First \texttt{(M)}} & {96.75}     & {95.87}  \\
\midrule

\rowcolor{gray!20}
\textbf{Bidirectional \texttt{(D)}} & \textbf{96.82}    & \textbf{99.66}    \\
\rowcolor{gray!5}
{w/o Table First \texttt{(D)}} & {95.89}     & {99.66}    \\
\rowcolor{gray!5}
{w/o Column First \texttt{(D)}} & {96.47}     & {98.70}     \\
\midrule

\rowcolor{gray!20}
\textbf{Bidirectional \texttt{(Q)}} & \textbf{94.88}     & \textbf{91.67}      \\
\rowcolor{gray!5}
{w/o Table First \texttt{(Q)}} & {88.15}     & {90.72}   \\
\rowcolor{gray!5}
{w/o Column First \texttt{(Q)}} & {90.46}     & {91.67}    \\

\bottomrule
\end{tabular}
\caption{Table Retrieval Performance on BIRD and Spider datasets.}
\label{tab:table_recall}
\end{table}


\begin{table*}[ht]
\centering
\small
\setlength{\tabcolsep}{8pt}
\renewcommand{\arraystretch}{1.2}
\begin{tabular}{%
    >{\raggedright\arraybackslash}p{4.3cm} |
    >{\raggedright\arraybackslash}p{9cm} |
    S[table-format=2.2]
}
\toprule
\rowcolor{red!10}
\textbf{Type} & \textbf{Description} & \textbf{(\%)} \\
\midrule
Missed Explicit Column & Column name is directly referred to in the question is not linked due to model oversight or prioritization issues. & {37} \\
\midrule
Name Mismatch & The retrieved schema fails to include required columns due to variations in naming, formatting, abbreviations, or mismatched casing. & {33} \\
\midrule
Partial Match Missed & Column partially matches tokens in the question but is missed due to insufficient lexical or semantic matching. & {23} \\
\midrule
Missed Implicit Table Context & Table needed to answer the question is not explicitly mentioned or hinted at in the question text, leading to linking failure. & {3} \\
\midrule
Missed Implicit Column Context & Column is semantically relevant but not explicitly mentioned, requiring inference that the model fails to perform. & {2} \\
\midrule
Ambiguous Numerical Reference & Numbers mentioned in the question (e.g., years, percentages, counts) match multiple possible fields in the schema, creating confusion. & {2} \\
\bottomrule
\end{tabular}
\caption{Distribution of detailed schema linking error categories. Percentages reflect the proportion of total observed errors attributed to each specific type. These insights highlight common challenges in schema linking and guide future model improvements.}
\label{tab:refined_error_analysis}
\end{table*}




\paragraph{Table Retrieval Performance} Table~\ref{tab:table_recall} presents the table retrieval performance of our bidirectional schema linking approach and its ablations on the BIRD and Spider datasets. Across all evaluated models, the full bidirectional strategy consistently achieves the highest recall on both datasets. Ablating either the table-first or column-first component leads to a noticeable drop in performance, particularly on the BIRD dataset, highlighting the complementary strengths of combining both retrieval directions.

\section{Error Analysis}
\label{appendix:error_analysis}

To better understand the limitations of our schema retrieval and SQL generation framework, we performed a detailed error analysis on cases where the retrieved schema failed to include gold schema elements on BIRD dataset. This qualitative analysis reveals several recurring patterns, which we categorize into six key error types.

The most prevalent issue, \textit{Missed Explicit Column}, occurs when the question explicitly refers to a column, yet the model fails to link it. This typically stems from incomplete keyword matching or failure to prioritize clearly referenced schema elements. Closely following is \textit{Schema Name Mismatch}, where column names are missed due to inconsistencies in formatting, abbreviation, or casing—highlighting the importance of robust name normalization and alias resolution.

Another significant category, \textit{Partial Match Missed}, involves cases where column names partially match question tokens but are still overlooked due to weak semantic generalization. These errors suggest opportunities for improved fuzzy matching or contextual embedding strategies. \textit{Missed Implicit Table Context} and \textit{Missed Implicit Column Context} reflect situations where the model fails to infer schema elements that are semantically required but not explicitly mentioned. These highlight the model’s limitations in performing multi-hop reasoning or leveraging database-specific priors.

Finally, \textit{Ambiguous Numerical Reference} captures cases where numeric values in the question (e.g., years, percentages, counts) could correspond to multiple candidate columns. Without additional disambiguation, the model often retrieves incorrect fields or conditions.

These error types reflect both lexical and semantic challenges in schema linking, and they motivate future work on enhanced disambiguation, context-aware retrieval, and reasoning-aware schema grounding. Table~\ref{tab:refined_error_analysis} summarizes these error categories and their relative frequencies and some examples are discussed in Appendix~\ref{appendix:error_example_case}. 


\begin{table*}[h]
\centering
\small
\setlength{\tabcolsep}{6pt}
\begin{tabular}{l|l|c}
\toprule
\textbf{Aspect Evaluated} & \textbf{Notes} & \textbf{\% of Samples} \\
\midrule
Subquestion usefulness & Logically decomposed and semantically accurate & 82\% \\
\midrule
Subquestion redundancy or vagueness & Repetition or generic rewording & 14\% \\
\midrule
Unhelpful subquestions & No useful decomposition provided & 4\% \\
\midrule
\midrule
Keyword relevance & Schema-aligned and discriminative & 81\% \\
\midrule
Generic or noisy keywords & Common verbs or non-informative terms & 14\% \\
\midrule
Hallucinated/irrelevant keywords & Terms not present in schema & 5\% \\
\bottomrule
\end{tabular}
\caption{Manual evaluation of subquestion and keyword quality (100 examples).}
\label{tab:error-eval_qaug}
\end{table*}


\paragraph{Errors in Question Augmentation.}
We also conducted a manual evaluation on 100 randomly sampled examples from the BIRD dataset to assess the quality of subquestions and extracted keywords generated by our enhancement module. The results, summarized in Table \ref{tab:error-eval_qaug}, indicate that approximately 82\% of subquestions were logically decomposed and semantically accurate, while 81\% of the keywords were aligned with schema elements. Nonetheless, a small fraction of examples exhibited redundancy, vagueness, or hallucinations, often due to overly generic keywords or weak subquestion structure. We have incorporated a discussion of these observations in the revised version of the paper and proposed improvements such as keyword filtering and fallback strategies. This evaluation underscores the robustness of our augmentation pipeline while highlighting directions for future refinement.


\section{Implementation Details}
\label{appendix:implementation_detail}
For our implementation, we utilize state-of-the-art large language models to perform schema linking and generate intermediate outputs for the Text-to-SQL pipeline. Specifically, we experiment with four large language models: \texttt{gpt-4o-mini}, \texttt{DeepSeek-chat}, \texttt{Qwen-2.5-7B-Instruct} and \texttt{gemini-2.0-flash}. These models are selected for their strong language understanding capabilities and efficient inference speed, making them well-suited for structured data reasoning tasks.

We interact with the models via their respective APIs and employ a temperature setting of 0.3 during inference to ensure output consistency and minimize randomness. This controlled decoding setting is crucial for tasks like schema linking, where precision and determinism are important. 

To rigorously assess the effectiveness of our method, we deliberately kept the prompt design minimal. This approach allows us to isolate and highlight the true impact of our schema linking strategy, independent of prompt engineering optimizations. In contrast, many recent models report higher performance in part due to extensive prompt engineering or tuning, which can confound the evaluation of the core method itself. By minimizing prompt interventions, we present a more direct measure of our method’s contribution. All prompts are carefully designed using zero-shot formats, and depending on the task we ensure that schema metadata and question context are passed in a compact and structured format (json) to maximize effectiveness within model context length limits. The source code for the implementation discussed in this paper will be released upon paper acceptance. 

\paragraph{Handling Sensitive Tokens in LLM Prompts}

While using \texttt{Gemini 2.0-flash}, we found that the model often refused to generate responses when prompts included sensitive tokens such as URLs or website links—typically citing safety or content policy concerns. This presented challenges in schema linking, especially when sample column values contained web addresses. To resolve this, we implemented a sanitization strategy to automatically remove such tokens from sample values before prompting, ensuring smooth generation and maintaining semantic intent. This experience highlights the importance of robust preprocessing when applying LLMs to structured data containing sensitive or user-generated content, and suggests future work on making LLMs more resilient to benign but structurally sensitive inputs like URLs, timestamps, or identifiers.

\paragraph{Formulation and Algorithm}

Let \( Q \) denote a natural language question over a set of table \( T \), and let AQ represent the augmented version of the question, as described in Section~\ref{sec:QuestioAugmentation}. Given a database schema 
\[ 
S = \{(t, c) | t \in T, c \in \text{Att}(t)\} 
\]
where $Att(t)$ denotes the set of attributes (columns) of table $t$, our objective is to identify a subset \( S^* \subseteq S \) containing the minimal set of relevant tables and columns necessary to generate a correct SQL query for $Q$. 

To achieve this, we employ a table-first retrieval strategy to select a subset of tables $T^*$ that are deemed relevant. In parallel, we use a column-first retrieval strategy to identify relevant columns $C^* \subseteq S$. The final retrieved schema $S^*$ is then defined as:
\[
\{(t, c) | t \in T^*, c \in \text{Att}(t)\} \cup C^*.
\]

The overall steps are summarized in Algorithm~\ref{alg:bidirectional-schema-linking}.



\begin{algorithm}[ht]
\caption{Bidirectional Schema Linking}
\label{alg:bidirectional-schema-linking}
\begin{algorithmic}[1]
\Require Natural language question \( Q \), database schema \( S = \{(t, c) \mid t \in T,\ c \in \text{Att}(t)\} \), LLM model
\Ensure Retrieved schema \( S^* \subseteq S \)

\State \textbf{Step 1: Question Augmentation}
\State \( AQ \gets \text{ExtractAugmentation}(Q, S) \) \Comment{Extract keywords, key phrases, subquestions}

\State \textbf{Step 2: Table-first schema linking}
\State \( T^* \gets \text{TableFirstFilter}(Q, AQ, S) \)

\State \textbf{Step 3: Column-first schema linking}
\State \( C^* \gets \text{ColumnFirstFilter}(Q, AQ, S) \)

\State \textbf{Step 4: Merge schemas}
\State \( S^* \gets \text{MergeSchemas}(T^*, C^*) \)

\State \Return \( S^* \)
\end{algorithmic}
\end{algorithm}

\section{Design Rationale and Justification}

To enhance schema linking in large-scale databases, we decompose the linking process into three complementary strategies: Table‑First, Column‑First, and Union‑based Schema Merging. Below, we provide a detailed, critical rationale for each choice.

\paragraph{Q1. Why use the Table‑First approach for schema linking?}  
The Table-First strategy is motivated by the observation that natural language questions often contain high-level entity or concept references that map more directly to table names. For instance, in a question like ``List the titles of books published by HarperCollins,'' the keyword ``books'' strongly hints at the \texttt{Books} table. By initially filtering for relevant tables, we reduce the schema space early and limit the search space for column-level linking in the next stage. This approach is particularly helpful when the schema is wide (many columns) but the number of tables is manageable. It ensures contextual precision and reduces unnecessary column comparisons, improving overall efficiency.

Natural language queries frequently include broad entity references that align directly with table names (e.g., “books,” “customers,” “orders”). By first filtering at the table level, we drastically reduce the candidate schema space before finer-grained column selection. This early pruning yields several benefits: (1) \emph{Token Efficiency}: fewer tables mean fewer tokens in subsequent prompts, mitigating context‑window constraints of LLMs; (2) \emph{Disambiguation}: coarse‑level entity matching leverages strong lexical signals in table names, reducing the risk of mis‑linking columns from irrelevant tables; and (3) \emph{Computational Savings}: limiting downstream column selection to a small subset of tables lowers the number of LLM calls and token processing overhead. In essence, Table‑First acts as a high‑precision filter that sets the stage for more reliable column retrieval, particularly in schemas with hundreds of tables but only a few relevant to the query.

\paragraph{Q2. Why choose the Column‑First approach for schema linking?} 

The Column-First method is chosen based on the insight that user questions frequently refer to fine-grained attributes rather than entire entities. For example, in a question like ``Which author has the highest average rating?'', the terms ``author'' and ``average rating'' point to specific columns across potentially multiple tables. In such cases, identifying relevant columns first allows us to backtrack and infer which tables they belong to. This is especially beneficial when column names carry meaningful semantics and table names are generic or ambiguous. It also allows the system to exploit strong lexical and semantic signals present in column headers for accurate linking.

In contrast, many NL questions hinge on attribute‑level clues rather than table entities (e.g., “highest average rating,” “total sales in Q4”). Column names often carry richer semantic content—numerical aggregations, timestamps, or descriptive fields—that can directly pinpoint the intended query targets. By identifying relevant columns first, we exploit these fine‑grained signals to (1) \emph{Recover Implicit Tables}: columns may indicate relationships or join paths that are not explicitly mentioned in the question; (2) \emph{Resolve Ambiguities}: when table names are generic or overlapping (e.g., “data\_log” vs. “event\_log”), column headers provide a stronger basis for disambiguation; and (3) \emph{Improve Recall for Attribute‑Driven Queries}: queries focused on specific metrics or filters benefit from attribute‑centric retrieval, ensuring that important but implicitly referenced tables are not overlooked. This bottom‑up approach balances the top‑down Table‑First strategy, capturing cases where column semantics lead the way.

\paragraph{Q3. What are the advantages of these approaches?}  
Both strategies provide complementary strengths. The Table-First approach excels in coarse-level disambiguation and reduces the overall schema context, making it more suitable for queries involving broad entities. The Column-First approach, on the other hand, is advantageous for queries with detailed attribute-level information, allowing finer granularity in linking. Additionally, these modular strategies allow independent error analysis and optimization at each stage. By decoupling the linking process, we improve interpretability and robustness, and we make the system more generalizable across datasets with varying schema complexities.

Together, Table‑First and Column‑First offer complementary strengths. The Table‑First strategy excels at high‑level pruning and reducing schema complexity, while Column‑First shines in capturing detailed, attribute‑driven requirements. Decomposing schema linking into these modular phases affords several advantages: (1) \emph{Error Isolation}: faults in table filtering do not propagate unchecked to column selection, and vice versa; (2) \emph{Targeted Optimization}: each phase can be tuned or augmented independently (e.g., specialized prompts for column disambiguation); (3) \emph{Interpretability}: analysts can inspect intermediate outputs to diagnose linking failures; and (4) \emph{Generalizability}: the same framework adapts to diverse schemas, from narrow tables with many columns to wide schemas with few tables. This modularity renders our method robust across varying database designs and query patterns.

\paragraph{Q4. Why take the Union for schema merging?}  
After obtaining table and column candidates from both directions, we apply a union-based merging strategy to form the final reduced schema. This design is grounded in the principle of recall maximization—each strategy may recover different subsets of relevant schema elements, and merging their outputs increases the likelihood of retaining all necessary components for correct SQL generation. For instance, in complex queries involving implicit joins or indirect references, one strategy may detect a critical table or column missed by the other. By taking the union, we combine their strengths while still achieving substantial schema reduction compared to using the full schema. This simple yet effective merging ensures that no essential part of the schema is omitted due to the limitations of a single perspective.

Each retrieval direction may uncover relevant schema elements missed by the other—Table‑First might omit columns whose table context is unclear, and Column‑First might miss tables whose columns lack distinctive values. By performing a union of both sets, we (1) \emph{Maximize Recall}: ensure that no critical table or column is excluded, approximating the perfect schema; (2) \emph{Maintain Precision}: because each direction already applies focused filtering, their union remains significantly smaller than the full schema; and (3) \emph{Simplify Integration}: a union operation is straightforward to implement and incurs negligible extra cost compared to re‑invoking the LLM. This union‑based merging thus captures the strengths of both strategies, providing a comprehensive yet compact schema that drives high‑accuracy SQL generation without downstream correction.

Together, these design decisions form a coherent, LLM‑friendly schema linking framework that narrows the performance gap between full and perfect schema usage, while balancing recall, precision, and computational efficiency.  

\section{Extended Related Work}

\paragraph{Text-to-SQL Pipelines}

Earlier Text-to-SQL models adopted modular pipelines, separating schema linking, parsing, and SQL generation. For example, IRNet used intermediate representations (SemQL) to bridge NL queries and SQL forms \cite{liu2024surveynl2sqllargelanguage}. Lei et al. \cite{lei-etal-2020-examining} highlighted the crucial role of schema linking by providing a labeled corpus and showing that performance heavily depends on accurate linking.

Several works enhanced traditional pipelines through query decomposition and reasoning. DIN-SQL \cite{pourreza2024din} introduced self-correction and step-by-step reasoning with decomposed in-context learning. CHASE-SQL \cite{pourreza2024chasesqlmultipathreasoningpreference} further extended this idea by combining multi-path reasoning and candidate reranking.

Schema linking is a fundamental step in the Text-to-SQL pipeline, enabling alignment between the natural language question and relevant database elements. Traditional approaches used string matching or hand-crafted rules, but recent methods utilize learned, context-aware linking strategies. \citet{glass2025extractive} introduce an extractive schema linking mechanism designed for decoder-only LLMs. Their method selects the most relevant schema components using a precision-oriented objective and demonstrates strong performance improvements on the Spider dataset, especially with lightweight LLMs.

\citet{cao2024rsl} propose RSL-SQL, a robust schema linking framework that combines bidirectional linking, context enhancement, and a binary selection strategy. This approach improves both precision and recall in schema linking, and includes a multi-turn self-correction mechanism that filters irrelevant elements during inference. RSL-SQL sets new benchmarks on BIRD and Spider datasets. \citet{10.1007/978-3-031-68309-1_11} develop a SQL-to-schema technique that first produces a preliminary SQL query using the full schema and then extracts only the referenced tables and columns to create a reduced schema. This method enables faster inference and better generalization for small models like CodeLlama-7B.

\citet{yuan2025knapsack} formulate schema linking as a knapsack optimization problem in their KaSLA framework. Using dynamic programming, KaSLA selects schema items by maximizing a relevance-utility score while respecting a context window budget. Their hierarchical design and use of binary linking classifiers allow it to scale to large databases. \citet{wang2025linkalign} introduce LinkAlign, a multi-phase schema linking approach for multi-database settings. It uses multi-round retrieval with context-aware ranking and enhances retrieved schemas using column-level semantics and query decomposition, significantly improving linking performance in real-world applications like BIRD.

\paragraph{Reasoning and Robustness in Text-to-SQL.}
In addition to schema linking, several works aim to improve the reasoning capabilities of LLMs in Text-to-SQL generation. \citet{excot2025} propose ExCoT, a framework that combines Chain-of-Thought (CoT) reasoning with off-policy and on-policy direct preference optimization (DPO). Rather than relying on ground truth SQL queries, ExCoT uses only execution feedback to fine-tune models, enabling performance improvements in weak supervision setups.

\citet{dai2024readsql} present READ-SQL, a reasoning path decomposer that breaks SQL queries into clauses and sub-SQLs. The READER module is used for both training data construction and post-hoc self-correction, improving confidence estimation and clause-level accuracy. \citet{stoisser2025sparks} introduce a reinforcement learning approach for training LLMs to reason over tables via SQL supervision. Their two-phase method first trains models to mimic SQL queries and then fine-tunes them to maximize execution correctness, showing that SQL generation can serve as a proxy for teaching tabular reasoning.


Approaches like E-SQL \cite{caferoğlu2025esqldirectschemalinking} enrich questions with schema context to enable direct alignment, while RSL-SQL \cite{cao2024rsl} utilizes partial query execution to validate links and improve robustness.

However, recent studies challenge the necessity of explicit linking. For instance, Maamari et al. \cite{maamari2024the} question whether schema linking is still needed when using well-reasoned LLMs, suggesting that modern models can infer the correct schema context from raw text. Despite this, practical evaluations (e.g., on Spider or BIRD) show that omitting schema linking can lead to performance drops due to hallucinations or misinterpretation in complex schemas \cite{li2024can}.

Our work is complementary: we treat schema linking as a standalone problem and evaluate its downstream impact on SQL generation under zero-shot settings.


The advent of LLMs shifted the paradigm toward in-context and retrieval-based prompting\cite{lee-etal-2025-mcs, liu-etal-2025-solid, nahid2025prism}. MCS-SQL \cite{lee-etal-2025-mcs} shows that using multiple prompts and multiple-choice ranking significantly improves accuracy by broadening the SQL generation search space. Solid-SQL \cite{liu2024solidsqlenhancedschemalinkingbased} combines schema-aware prompting with retrieval strategies to improve robustness.

CHESS \cite{talaei2024chesscontextualharnessingefficient} represents a multi-agent approach that incrementally refines prompts, demonstrating improved linking and execution performance but at the cost of many LLM calls. MAC-SQL \cite{wang2024macsqlmultiagentcollaborativeframework} follows a similar path with a collaborative framework involving multiple agents that reason jointly over query elements. Our method in this work builds on these insights but introduces a more structured bidirectional schema linking framework that balances precision and recall.

\paragraph{Benchmarks: BIRD and Spider.}
Two key benchmarks have driven progress in the field.

\citet{yu-etal-2018-spider} introduce Spider, a large-scale Text-to-SQL dataset that spans 200 databases across various domains. It is particularly challenging due to its use of unseen schemas in the test set, requiring models to generalize beyond memorization.

\citet{NEURIPS2023_83fc8fab} develop BIRD, a benchmark tailored for large-scale, real-world database Text-to-SQL generation. The dataset includes over 12,000 examples with 95 large databases, noisy text fields, and long context windows. BIRD emphasizes schema selection, ambiguity resolution, and execution grounding, providing a comprehensive challenge for modern LLMs.

\paragraph{Surveys and Analyses.}
Several recent surveys have summarized the field’s evolution.

\citet{zhu2024llm, deng-etal-2022-recent} provide an extensive review of LLM-enhanced Text-to-SQL systems, categorizing them into prompt-based, fine-tuned, pre-trained, and agent-based models. They also summarize common datasets, evaluation strategies, and prompting frameworks used in the field.

Recent comprehensive survey by~\citet{hong2025nextgenerationdatabaseinterfacessurvey} provides an in-depth overview of the rapidly evolving landscape of LLM-based text-to-SQL systems. The authors systematically examine next-generation database interfaces, covering advances in model architectures, training strategies, prompt engineering, and real-world applications. This survey offers valuable context for ongoing research, highlighting key challenges and future directions in bridging natural language understanding and structured data querying.

\paragraph{Applications.}

The ability of natural language interfaces to interact with structured data is central to developing robust and interpretable reasoning systems. Text-to-SQL systems, in particular, not only streamline querying of relational databases but also enable a wide range of structured data reasoning applications. Recent works have leveraged text-to-SQL for tasks such as table-based question answering, compositional reasoning, and data interpretation across diverse domains~\cite{abhyankar-etal-2025-h, nahid-rafiei-2024-tabsqlify, zhang2023reactable, nahid-rafiei-2024-normtab, deng-etal-2022-recent}. These advances highlight the growing versatility and impact of text-to-SQL frameworks in supporting both practical and research-oriented data analysis scenarios.

\onecolumn

\section{Schema Linking Examples}
\label{appendix:illustrative_example_cases}
We present three illustrative outputs from our schema linking framework to demonstrate how the method identifies relevant tables and columns for a given question (see Figure~\ref{fig:exampe1}, \ref{fig:exampe2} and \ref{fig:exampe3}). The examples showcase both strengths and challenges of our approach in aligning natural language with database schema. Each example includes question augmentation, gold schema, and outputs from both retrieval directions.


\begin{figure}[ht]
\centering
\normalsize
\begin{tcolorbox}[colframe=teal!70!black, colback=teal!5!white, coltitle=white, fonttitle=\bfseries, fontupper=\itshape, title=\textbf{Example 1: Effective Schema Retrieval Using Table-First Approach}]

\textbf{Question-ID:} \texttt{1297} \\ 
\textbf{DB-ID:} \texttt{thrombosis\_prediction} \\ 
\textbf{Difficulty:} \texttt{moderate} \\ 
\noindent\rule{\textwidth}{0.4pt}
\textbf{Gold-Schema:}  \texttt{\{'Examination': ['ID', 'KCT'], 'Patient': ['ID'], 'Laboratory': ['ID', 'T-CHO']\}}\\
\noindent\rule{\textwidth}{0.4pt}

\textbf{Question Augmentation:}  \\

\textbf{Original Question:} \texttt{For the patients whose total cholesterol is higher than normal, how many of them have a negative measure of degree of coagulation?}\\

\textbf{Sub-questions:} \texttt{["What is the normal range for total cholesterol?", "Which patients have total cholesterol higher than the normal range?", "What is the measure of degree of coagulation for each of those patients?", "How many of those patients have a negative measure of degree of coagulation?"]} \\

\textbf{keywords and key-phrases:} \texttt{['patients', 'total cholesterol', 'higher than normal', 'negative measure', 'degree of coagulation', 'T-CHO', '>= 250', 'KCT', '-']} \\
\noindent\rule{\textwidth}{0.4pt}

\textbf{Tables First Schema:} \\
\texttt{\{'Laboratory': ['ID', 'T-CHO'], 'Examination': ['ID', 'KCT'], \textcolor{blue}{\textbf{'Patient': ['ID']}}\}} \\ 
 \noindent\rule{\textwidth}{0.4pt}
\textbf{Column First Schema:} \\
\texttt{\{'Laboratory': ['ID', 'T-CHO'], 'Examination': ['ID', 'KCT']\}} \\ 
 \noindent\rule{\textwidth}{0.4pt}
\textbf{Bidirectional Schema:} \\
\texttt{\{'Laboratory': ['ID', 'T-CHO'], 'Examination': ['ID', 'KCT'], \textcolor{blue}{\textbf{'Patient': ['ID']}}\}} \\ 
\noindent\rule{\textwidth}{0.4pt}
\end{tcolorbox}
\caption{In this example, the table-first strategy successfully retrieves the \texttt{Patient} table based on high-level query understanding, while the column-first direction overlooks it due to missing low-level column clues. The bidirectional approach resolves this by unifying both perspectives.}
\label{fig:exampe1}
\end{figure}



\begin{figure}[ht]
\centering
\normalsize
\begin{tcolorbox}[colframe=teal!70!black, colback=teal!5!white, coltitle=white, fonttitle=\bfseries, fontupper=\itshape, title=\textbf{Example 2: Both Table First and Column First contributed to extract correct schema}]

\textbf{Question-ID:} \texttt{271} \\ 
\textbf{DB-ID:} \texttt{toxicology} \\ 
\textbf{Difficulty:} \texttt{simple} \\ 
\noindent\rule{\textwidth}{0.4pt}
\textbf{Gold-Schema:}  \texttt{\{'connected': ['bond\_id', 'atom\_id2', 'atom\_id'], 'atom': ['element', 'atom\_id']\}}\\
\noindent\rule{\textwidth}{0.4pt}

\textbf{Question Augmentation:}  \\

\textbf{Original Question:} \texttt{Does bond id TR001\_1\_8 have both element of chlorine and carbon?} \\

\textbf{Sub-questions:}  \texttt{["What elements does bond id TR001\_1\_8 have?",
"Does the list of elements for bond id TR001\_1\_8 include chlorine?",
"Does the list of elements for bond id TR001\_1\_8 include carbon?"]} \\

\textbf{keywords and key-phrases:} \texttt{["bond id", "TR001\_1\_8", "chlorine", 
"carbon", "element", "cl", "c"]} \\
\noindent\rule{\textwidth}{0.4pt}
\textbf{Tables First Schema:} \\
\begin{flushleft}
\texttt{\{}\\
    \hspace*{1em}\texttt{'Atom': ['atom\_id', 'element'],} \\
    \hspace*{1em}\texttt{'Bond': ['bond\_id'],} \\
    \hspace*{1em}\texttt{\textcolor{blue}{\textbf{'Connected': ['bond\_id', 'atom\_id', 'atom\_id2']}}} \\
\texttt{\}}
\end{flushleft}
\noindent\rule{\textwidth}{0.4pt}
\textbf{Column First Schema:} \\
\begin{flushleft}
\texttt{\{}\\
    \hspace*{1em}\texttt{'Atom': ['atom\_id', \textcolor{blue}{\textbf{'molecule\_id'}}, 'element'],} \\
    \hspace*{1em}\texttt{\textcolor{magenta}{'Bond': ['bond\_id', 'molecule\_id'}]} \\
\texttt{\}} 
\end{flushleft}
\noindent\rule{\textwidth}{0.4pt}
\textbf{Bidirectional Schema:} \\
\begin{flushleft}
\texttt{\{}\\
\hspace*{1em}\texttt{\textcolor{blue}{'Atom'}: [\textcolor{blue}{\textbf{'molecule\_id'}}, 'element', 'atom\_id'],}\\
\hspace*{1em}\texttt{\textcolor{magenta}{'Bond': ['bond\_id', 'molecule\_id']},}\\
\hspace*{1em}\texttt{\textcolor{blue}{\textbf{'Connected': ['bond\_id', 'atom\_id', 'atom\_id2']}}}\\
\texttt{\}}
\end{flushleft}

\noindent\rule{\textwidth}{0.4pt}
\end{tcolorbox}
\caption{Example illustrating how both table-first and column-first strategies contribute to accurate schema retrieval. For Question-ID 271 from the \texttt{toxicology} database, elements from both strategies complement each other to form the correct bidirectional schema, closely matching the gold schema and enabling precise SQL generation.}

\label{fig:exampe2}
\end{figure} 

\begin{figure}[ht]
\centering
\normalsize
\begin{tcolorbox}[colframe=teal!70!black, colback=teal!5!white, coltitle=white, fonttitle=\bfseries, fontupper=\itshape, title=\textbf{Example 3: Column-First Advantage in Recovering Implicit Identifier Columns}]

\textbf{Question-ID:} \texttt{291} \\ 
\textbf{DB-ID:} \texttt{toxicology} \\ 
\textbf{Difficulty:} \texttt{simple} \\ 
\noindent\rule{\textwidth}{0.4pt}
\textbf{Gold-Schema:}  \texttt{\{'molecule': ['molecule\_id', 'label']\}}\\
\noindent\rule{\textwidth}{0.4pt}

\textbf{Question Augmentation:}  \\

\textbf{Original Question:} \texttt{How many chemical compounds in the database are identified as carcinogenic.} \\

\textbf{Sub-questions:}  \texttt{["What are the chemical compounds in the database?", "Which of those compounds are identified as carcinogenic?", "How many compounds are identified as carcinogenic?"]} \\

\textbf{keywords and key-phrases:} \texttt{["chemical compounds", "database", "carcinogenic", "label", "+", "molecules", "carcinogenic"]} \\
\noindent\rule{\textwidth}{0.4pt}
\textbf{Tables First Schema:} \texttt{\{'Molecule': ['label']\}} \\ 
\noindent\rule{\textwidth}{0.4pt}
\textbf{Column First Schema:} \texttt{\{'Molecule': [\textcolor{blue}{\textbf{'molecule\_id'}}, 'label']\}} \\ 
\noindent\rule{\textwidth}{0.4pt}
\textbf{Bidirectional Schema:} \texttt{\{'Molecule': ['label',\textcolor{blue}{\textbf{'molecule\_id'}}]\}} \\ 
\noindent\rule{\textwidth}{0.4pt}
\end{tcolorbox}
\caption{In this example, the column-first strategy successfully includes the identifier column \texttt{molecule\_id}, which the table-first strategy misses. This highlights how focusing on fine-grained column relevance can capture crucial schema elements for accurate SQL construction.}

\label{fig:exampe3}
\end{figure} 

\clearpage

\section{Prompt Templates}
\label{appendix:appendix-prompts}
In this section, we present the prompt templates used in our schema linking and SQL generation framework. These templates were carefully designed to guide large language models (LLMs) in performing key tasks such as question augmentation, table and column retrieval, and final SQL query generation. By providing clear instructions, structured input formats, and examples, the prompts help ensure consistency, high recall, and interpretability across different components of our system. We include both the schema retrieval and SQL generation prompt designs to support reproducibility and future research.

\begin{figure}[ht]
\centering

\begin{tcolorbox}[colframe=blue!60!black, colback=blue!5!white, coltitle=white, fonttitle=\bfseries, fontupper=\itshape, title=\textbf{Column Retrieval Prompt}]
\normalsize 
You are tasked with identifying the relevant columns from a database schema needed to write an SQL query form a natural language question. The schema is provided as a dictionary where keys are table names and values are lists of column names (e.g., \{"Employees": ["ID", "Name"]\}). You will receive: \\
- The original question \\ 
- A list of subquestions breaking down the original question into simpler parts \\ 
- A list of keywords and keyphrases extracted from the question \\ 
- The schema containing tables and their full column lists \\ 
- An optional hint (if provided) \\
Your goal is to select the columns from each table required to construct an SQL query that answers the question. To ensure high recall, include any column that might be relevant based on the original question, subquestions, or keywords, such as columns for filtering, joining, or computing results. \\
Output the result as a JSON-formatted dictionary where keys are the table names from the filtered schema and values are the selected column names, like \{"table1": ["col1"], "table2": ["col2"]\}. \\

Here are some examples: \\
Example 1: [EXAMPLE] \\ 
Example 2: [EXAMPLE] \\ 
Example 3: [EXAMPLE] \\

Now, identify the relevant columns for the following: \\
Schema: \{SCHEMA\} \\
Question Information: \{AUGMENTED\_QUESTION\} \\
Hint: \{HINT\} \\

Please respond with a JSON object structured as follows: \\
\texttt{\{ \\
"chain of thought reasoning": "Your reasoning for selecting the columns, be concise and clear.", \\
"table name1": ["column1", "column2", ...], \\
"table name2": ["column1", "column2", ...], \\
...\\\}} \\
Make sure your response includes the table names as keys, each associated with a list of column names that are necessary for writing a SQL query to answer the question. Be cautious about the foreign keys. \\
For each aspect of the question, provide a clear and concise explanation of your reasoning behind selecting the columns. Do not include ````json in your response. Only output a json as your response.
\end{tcolorbox}
\caption{Schema Retrieval Prompt Template used for identifying relevant schema elements based on a given natural language question and database schema.}
\label{fig:column-retrieval-prompt}
\end{figure}

\begin{figure}[ht]
\centering

\begin{tcolorbox}[colframe=blue!60!black, colback=blue!5!white, coltitle=white, fonttitle=\bfseries, fontupper=\itshape, title=\textbf{Table Retrieval Prompt}]
\normalsize
You are tasked with identifying the relevant tables and all their columns from a database schema needed to write an SQL query from a natural language question. The schema is provided as a dictionary where keys are table names and values are lists of column names. You will receive: \\
- The original question \\ 
- A list of subquestions breaking down the original question into simpler parts \\ 
- A list of keywords and keyphrases extracted from the question \\ 
- The schema containing tables and their full column lists \\ 
- An optional hint (if provided) \\

Your goal is to select the tables required to construct an SQL query that answers the question and include all columns from those tables. To ensure high recall, include any table that might be relevant based on the original question, subquestions, keywords, or hint, even if its relevance is uncertain. \\
Output the result as a JSON-formatted dictionary where keys are the selected table names and values are their full column lists, like \{"table1": ["col1"], "table2": ["col2"]\}. \\

Here are some examples: \\
Example 1: [EXAMPLE] \\ 
Example 2: [EXAMPLE] \\ 
Example 3: [EXAMPLE] \\

Now, identify the relevant columns for the following: \\
Schema: \{SCHEMA\} \\
Question Information: \{AUGMENTED\_QUESTION\} \\
Hint: \{HINT\} \\

Please respond with a JSON object structured as follows: \\
\texttt{\{ \\
"chain of thought reasoning": "Your reasoning for selecting the columns, be concise and clear.", \\
"table name1": ["column1", "column2", ...], \\
"table name2": ["column1", "column2", ...], \\
...\\\}} \\

Make sure your response includes the table names as keys, each associated with
a list of column names that are necessary for writing a SQL query to answer the original question. Be cautious about the foreign keys.  \\
For each aspect of the question, provide a clear and concise explanation of your reasoning behind selecting the columns. Do not include ````json in your response. Only output a json as your response.
\end{tcolorbox}
\caption{Schema Retrieval Prompt Template used for identifying relevant schema elements based on a given natural language question and database schema.}
\label{fig:table-retrieval-prompt}
\end{figure}

\begin{figure}[ht]
\centering
\begin{tcolorbox}[colframe=blue!60!black, colback=blue!5!white, coltitle=white, fonttitle=\bfseries, fontupper=\itshape, title=\textbf{SQL Generation Prompt Template}]
\normalsize
You are a data science expert. Below, you are presented with a database schema and a question. Your task is to read the schema, understand the question, and generate a valid SQLite query to answer the original question.\\[6pt]

Before generating the final SQL query, think step by step about how to write the query.\\[6pt]

\textbf{Database Schema:} \{SCHEMA\} \\[6pt]

This schema offers an in-depth description of the database’s architecture, detailing tables, columns, primary keys, foreign keys, and any pertinent information regarding relationships or constraints. Special attention should be given to the examples listed beside each column (if any), as they directly hint at which columns are relevant to our query.\\[6pt]

\textbf{Database admin instructions:}
\begin{itemize}
    \item[-] Only output the information explicitly asked in the question. If the question asks for a specific column, include only that column in the SELECT clause.
    \item[-] The predicted query should return all of the information asked in the question—nothing more, nothing less.
\end{itemize}

\textbf{Question Information:} \{AUGMENTED\_QUESTION\} \\[6pt]
\textbf{Hint:} \{EVIDENCE\} \\[6pt]

The question information, including subquestions and keywords, is designed to guide your focus toward the most relevant parts of the schema needed to answer the question effectively.\\[6pt]

Please respond with a JSON object structured as follows:

\texttt{\{ "SQL": "Your SQL query in a single string." \}}\\[6pt]

Priority should be given to columns that have been explicitly matched with examples relevant to the question’s context.\\[6pt]

Take a deep breath and think step by step to find the correct SQLite SQL query for the original question.
\end{tcolorbox}
\caption{SQL Generation Prompt Template used for composing executable queries from natural language questions, schema metadata, and auxiliary hints.}
\label{fig:sql-generation-prompt}
\end{figure}

\begin{figure}[ht]
\centering
\begin{tcolorbox}[colframe=blue!60!black, colback=blue!5!white, coltitle=white, fonttitle=\bfseries, fontupper=\itshape, title=\textbf{Question Decomposition Prompt Template}]
\normalsize
You are tasked with decomposing a natural language question into smaller subquestions to help generate an SQL query from a database. \\
Your goal is to break the question into simple, specific subquestions that together fully address the original question. Each subquestion should focus on a single aspect or step needed to answer the original question, such as identifying data, filtering, or calculating something.\\[6pt]

\textbf{Here are some examples:} \\[4pt]

\textbf{Example 1:} \\
Original Question: ``What is the total revenue from orders placed in 2023?'' \\
Subquestions:
\begin{enumerate}
    \item ``Which orders were placed in 2023?''
    \item ``What is the revenue for each of those orders?''
    \item ``What is the total of that revenue?''
\end{enumerate}

\textbf{Example 2:} \\
Original Question: ``Which employees work in departments located in New York?'' \\
Subquestions:
\begin{enumerate}
    \item ``Which departments are located in New York?''
    \item ``Which employees work in those departments?''
\end{enumerate}

Now, decompose the following question into subquestions. Provide the subquestions as a numbered list.\\[6pt]

\textbf{Original Question:} \{QUESTION\} \\[6pt]

Provide the subquestions in a json object without any explanation.
Please respond with a JSON object structured as follows: \\ 

\texttt{\{ \\
"Subquestions": list of subquestions.\\
\}} \\

Do not include \texttt{```json} in your response. Only output a json object as your response.

\end{tcolorbox}
\caption{Question Decomposition Prompt Template used to extract subquestions that help guide SQL generation from a natural language query.}
\label{fig:question-decomposition-prompt}
\end{figure}
\begin{figure}[ht]
\centering
\begin{tcolorbox}[colframe=blue!60!black, colback=blue!5!white, coltitle=white, fonttitle=\bfseries, fontupper=\itshape, title=\textbf{Keyword and Keyphrase Extraction Prompt Template}]
\normalsize
\textbf{Objective:} Analyze the given question and hint to identify and extract keywords, keyphrases, and named entities. These elements are crucial for understanding the core components of the inquiry and the guidance provided. The goal is to recognize and isolate significant terms and phrases that could be instrumental in formulating searches or queries related to the posed question.\\[6pt]

\textbf{Instructions:}
\begin{enumerate}
    \item \textbf{Read the Question Carefully:} Identify the main focus, named entities (e.g., organizations, locations), technical terms, and key concepts.
    \item \textbf{Analyze the Hint:} Extract keywords or phrases that provide clarity or direction toward answering the question.
    \item \textbf{List Keyphrases and Entities:} Combine findings from both the question and hint into a single Python list. Include:
    \begin{itemize}
        \item \textbf{Keywords:} Essential single words.
        \item \textbf{Keyphrases:} Multi-word terms or named entities.
    \end{itemize}
    Ensure all terms maintain the original phrasing or terminology used in the input.
\end{enumerate}

\textbf{Task:} \\[4pt]
Given the following question and hint, identify and list all relevant keywords, keyphrases, and named entities.\\[6pt]

\textbf{Question:} \{QUESTION\} \\[4pt]
\textbf{Hint:} \{EVIDENCE\} \\[6pt]

Please provide your findings as a json file, capturing the essence of both
the question and hint through the identified terms and phrases. \\

\texttt{\{ \\
    "keywords": list of keywords, keyphrases and entities. \\
\}} \\

Do not include \texttt{```json} in your response. Only output a json object as your response.
\end{tcolorbox}
\caption{Keyword and Keyphrase Extraction Prompt Template for identifying essential terms, keyphrases, and named entities from a question and hint to guide schema retrieval.}
\label{fig:keyword-extraction-prompt}
\end{figure}

\clearpage
\section{Representative Error Case}
\label{appendix:error_example_case}

We present some representative failure case where the model failed to retrieve relevant schema components from BIRD dataset.








\begin{tcolorbox}[colframe=red!75!black, colback=red!5!white, fonttitle=\bfseries, fontupper=\itshape, title=\textbf{Example: Schema Name Discrepancy (QID: 130)}]
\textbf{DB:} \texttt{financial} \\
\textbf{Question:} How many of the account holders in South Bohemia still do not own credit cards?

\textbf{Gold Schema:} \\
\texttt{'disp': ['account\_id', 'client\_id', 'type'], 'client': ['district\_id', 'client\_id'], 'district': ['district\_id', 'A3']}

\textbf{Retrieved Schema (Bidirectional):} \\
\texttt{'District': ['district\_id', 'A3'], 'Client': ['district\_id', 'client\_id'], 'Disp': ['account\_id', 'client\_id'], 'Account': ['account\_id', 'district\_id'], 'Card': ['disp\_id']}

\textbf{Gold SQL:}
\begin{verbatim}
SELECT COUNT(T3.account_id) 
FROM district AS T1 
JOIN client AS T2 ON T1.district_id = T2.district_id 
JOIN disp AS T3 ON T2.client_id = T3.client_id 
WHERE T1.A3 = 'south Bohemia' AND T3.type != 'OWNER'
\end{verbatim}

\textbf{Missing Schema Element(s):} \\
\texttt{disp.type}

\textbf{Discussion:} In this example, although the column \texttt{disp.type} is semantically relevant and essential for answering the question, the model fails to retrieve it due to such mismatches.
\end{tcolorbox}


\begin{tcolorbox}[colframe=red!75!black, colback=red!5!white, fonttitle=\bfseries, fontupper=\itshape, title=\textbf{Example: Missed Explicit Column Mention (QID: 71)}]
\textbf{DB:} \texttt{california\_schools} \\
\textbf{Question:} List the schools in Alameda County that are eligible for free meals. \\

\textbf{Gold Schema:} \\
\texttt{frpm.School Name, frpm.County Name, frpm.Free Meal Count (K-12), frpm.Enrollment (K-12)} \\ 

\textbf{Retrieved Schema (Bidirectional):} \\
\texttt{frpm.County Name, frpm.Enrollment (K-12), frpm.CDSCode} \\ 

\textbf{Gold SQL:}
\begin{verbatim}
SELECT School Name 
FROM frpm 
WHERE County Name = 'Alameda' AND 
      Free Meal Count (K-12) > 0
\end{verbatim}

\textbf{Missing Schema Element(s):} \\
\texttt{frpm.School Name, frpm.Free Meal Count (K-12)}

\textbf{Discussion:} Although the column \texttt{Free Meal Count (K-12)} is semantically aligned with the phrase “eligible for free meals,” the model failed to retrieve it. This reflects a breakdown in recognizing direct relevance from natural language cues.
\end{tcolorbox}

\vspace{-5pt}








\begin{tcolorbox}[colframe=red!75!black, colback=red!5!white, fonttitle=\bfseries, fontupper=\itshape, title=\textbf{Example: Schema Name Discrepancy (QID: 141)}]
\textbf{DB:} \texttt{financial} \\
\textbf{Question:} Which districts have transactions greater than \$10,000 in 1997? \\

\textbf{Gold Schema:} \\
\texttt{trans.date, trans.amount, account.account\_id, client.district\_id, district.A3} \\

\textbf{Retrieved Schema (Bidirectional):} \\
\texttt{trans.date, trans.amount, trans.balance, account.account\_id, client.district\_id} \\

\textbf{Gold SQL:}
\begin{verbatim}
SELECT DISTINCT T4.A3 
FROM trans AS T1 
JOIN account AS T2 ON T1.account_id = T2.account_id 
JOIN client AS T3 ON T2.account_id = T3.account_id 
JOIN district AS T4 ON T3.district_id = T4.district_id 
WHERE T1.date LIKE '1997%' AND T1.amount > 10000
\end{verbatim}

\textbf{Missing Schema Element(s):} \\
\texttt{district.A3}

\vspace{6pt}
\textbf{Discussion:} The model fails to retrieve column \texttt{district.A3} due to casing and contextual mismatch, despite it being critical to map district names.
\end{tcolorbox}










\begin{tcolorbox}[colframe=red!75!black, colback=red!5!white, fonttitle=\bfseries, fontupper=\itshape, title=\textbf{Example: Missed Implicit Table Context (QID: 711)}]
\textbf{DB:} \texttt{student\_club} 

\textbf{Question:} Which zip codes have the most active members? \\

\textbf{Gold Schema:} \\
\texttt{member.zip, attendance.link\_to\_member} \\

\textbf{Retrieved Schema (Bidirectional):} \\
\texttt{member.zip, member.position, member.link\_to\_major} \\

\textbf{Gold SQL:}
\begin{verbatim}
SELECT zip, COUNT(*) 
FROM attendance 
JOIN member ON attendance.link_to_member = member.member_id 
GROUP BY zip 
ORDER BY COUNT(*) DESC
\end{verbatim}

\textbf{Missing Schema Element(s):} \\
\texttt{attendance.link\_to\_member}

\vspace{6pt}
\textbf{Discussion:} The attendance table is crucial to link membership activity but was not retrieved, likely because it was not explicitly referenced in the question.
\end{tcolorbox}



\end{document}